\documentclass[11pt,a4paper]{article}
\usepackage[hyperref]{acl2020}
\usepackage{times}
\usepackage{latexsym}

\usepackage{microtype}

\aclfinalcopy % Uncomment this line for the final submission
\def\aclpaperid{1931} %  Enter the acl Paper ID here

\usepackage{amsmath}
\usepackage{amssymb}
\usepackage{booktabs}
\usepackage{graphicx}
\usepackage{hyperref}
\usepackage{relsize}
\usepackage{latexsym}
\usepackage{paralist}
\usepackage{times}
\usepackage{url}
\usepackage{xcolor}
\usepackage{xspace}
\usepackage{diagbox}
\usepackage{multirow}
\usepackage{makecell}
\usepackage{bigdelim}
\usepackage{mathtools}

\date{}

\newcommand\sect[1]{\S\ref{#1}}
\definecolor{darkgreen}{rgb}{0.05, 0.5, 0.06}
\newcommand\bert{\textsc{BERT}\xspace}
\newcommand\roberta{\textsc{RoBERTa}\xspace}
\newcommand\vampire{\textsc{VAMPIRE}\xspace}

\newcommand\daptfull{Domain-Adaptive Pretraining\xspace}
\newcommand\taptfull{Task-Adaptive Pretraining\xspace}
\newcommand\dapt{\textsc{dapt}\xspace}
\newcommand\tapt{\textsc{tapt}\xspace}
\newcommand\oracle{Curated-\textsc{tapt}\xspace}
\newcommand\random{\textsc{rand-tapt}\xspace}
\newcommand\knn[1]{$#1$\textsc{nn-tapt}\xspace}

\newcommand\rct{\textsc{RCT}\xspace}

\newcommand\chemprot{\textsc{ChemProt}\xspace}
\newcommand\arccite{\textsc{ACL-ARC}\xspace}
\newcommand\sciie{\textsc{SciERC}\xspace}
\newcommand\hp{\textsc{HyperPartisan}\xspace}
\newcommand\hpshort{\textsc{HyP.}\xspace}
\newcommand\ag{\textsc{AGNews}\xspace}
\newcommand\helpful{\textsc{Helpfulness}\xspace}
\newcommand\imdb{\textsc{IMDB}\xspace}

\newcommand\news{\textsc{News}\xspace}
\newcommand\med{\textsc{BioMed}\xspace}
\newcommand\cs{\textsc{CS}\xspace}
\newcommand\realnews{\textsc{RealNews}\xspace}
\newcommand\reviews{\textsc{Reviews}\xspace}
\newcommand\PT{\textsc{PT}\xspace}

\newcommand\amazon{\textsc{Amazon} reviews\xspace}
\newcommand\gorc{\textsc{S2ORC}\xspace}

\newcommand\wiki{\textsc{Wikipedia}\xspace}
\newcommand\books{\textsc{BookCorpus}\xspace}
\newcommand\ccnews{\textsc{CCNews}\xspace}

\title{Don't Stop Pretraining: Adapt Language Models to Domains and Tasks}

\author{
    Suchin Gururangan$^\dagger$ \quad
	Ana Marasovi\'{c}$^{\dagger\diamondsuit}$ \quad
	Swabha Swayamdipta$^{\dagger}$ \quad \\
	\bf Kyle Lo$^\dagger$ \quad
	Iz Beltagy$^\dagger$ \quad
	Doug Downey$^\dagger$ \quad Noah A. Smith$^{\dagger\diamondsuit}$ \\\\
	$^\dagger$Allen Institute for Artificial Intelligence, Seattle, WA, USA \\
	$^\diamondsuit$Paul G. Allen School of Computer Science \& Engineering, University of Washington, Seattle, WA, USA \\
	{\tt \{suching,anam,swabhas,kylel,beltagy,dougd,noah\}@allenai.org}
}

\begin{document}
\maketitle

\begin{abstract}
Language models pretrained on text from a wide variety of sources form the foundation of today's NLP. 
In light of the success of these broad-coverage models, we investigate whether it is still helpful to tailor a pretrained model to the domain of a target task.
We present a study across four domains (biomedical and computer science publications, news, and reviews) and eight classification tasks, showing that a second phase of pretraining in-domain (\textit{domain-adaptive pretraining}) leads to performance gains, under both high- and low-resource settings. Moreover, adapting to the task's unlabeled data (\textit{task-adaptive pretraining}) improves performance even after domain-adaptive pretraining.
Finally, we show that adapting to a task corpus augmented using simple data selection strategies is an effective alternative, especially when resources for domain-adaptive pretraining might be unavailable.
Overall, we consistently find that multi-phase adaptive pretraining offers large gains in task performance.
\end{abstract}
\section{Introduction}
\label{sec:intro}

\begin{figure}[t]
     \centering
     \includegraphics[width=3.0in]{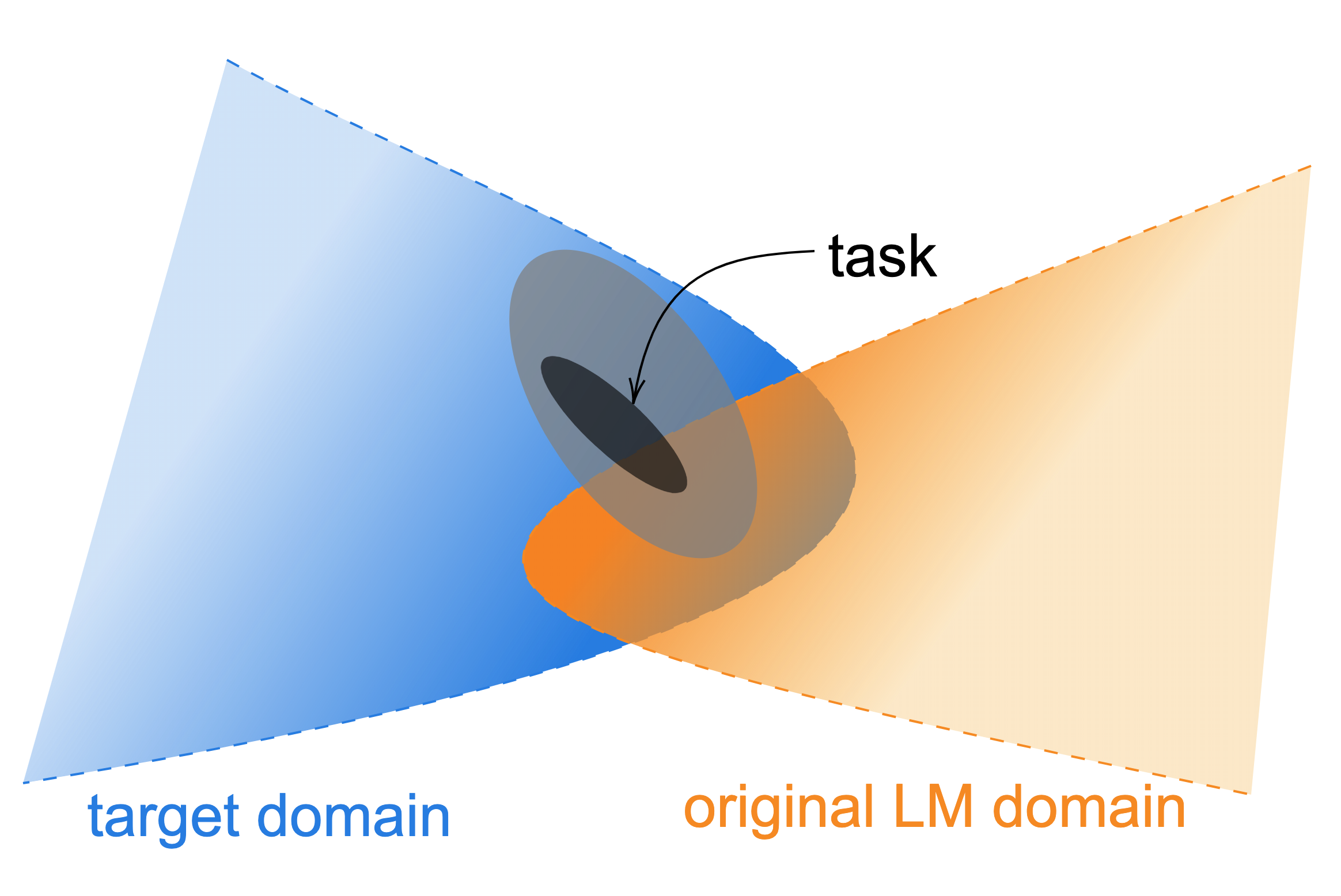}
     \caption{An illustration of data distributions.
     Task data is comprised of an observable task distribution, usually non-randomly sampled from a wider distribution (light grey ellipsis) within an even larger target domain, which is not necessarily one of the domains included in the original LM pretraining domain -- though overlap is possible.  We explore the benefits of continued pretraining on data from the task distribution and the domain distribution. 
     } 
     \label{fig:domains_intuition}
 \end{figure}
 
Today's pretrained language models are trained on massive, heterogeneous corpora \cite{Raffel2019ExploringTL, Yang2019XLNetGA}.  
For instance, \roberta \cite{Liu2019RoBERTaAR} was trained on over 160GB of uncompressed text, with sources ranging from English-language encyclopedic and news articles, to literary works and web content.
Representations learned by such models achieve strong performance across many tasks with datasets of varying sizes drawn from a variety of sources \citep[e.g., ][]{wang-etal-2018-glue,Wang2019SuperGLUEAS}.
This leads us to ask whether a task’s textual {\em domain}---a term typically used to denote a distribution over language characterizing a given topic or genre (such as ``science'' or ``mystery novels'')---is still relevant.
Do the latest large pretrained models work universally or is it still helpful to build separate pretrained models for specific domains?

While some studies have shown the benefit of continued pretraining on domain-specific unlabeled data \cite[e.g.,][]{Lee2019BioBERTAP}, %especially under a low-resource setting \cite[e.g.,][]{han_eisenstein_emnlp_2019},
these studies only consider a single domain at a time and use a language model that is pretrained on a smaller and less diverse corpus than the most recent language models.
% \ana{Revise?}
% whether continued pretraining is beneficial for today's strongest pretrained models,\ana{Mark: it is not clear that "the strongest pretrained models" here are those trained on diverse corpus such as RoBERTa's and hence it seemed to him like this was actually confirmed by SciBERT/BioBERT.} or 
Moreover, it is not known how the benefit of continued pretraining may vary with factors like the amount of available labeled task data, or the proximity of the target domain to the original pretraining corpus (see Figure \ref{fig:domains_intuition}).
%(e.g., since reviews are web content, they are kind of close to the OpenWebText Corpus \roberta is trained on).

We address this question %whether continued pretraining on target domains is beneficial
for one such high-performing model, \roberta \cite{Liu2019RoBERTaAR} (\sect{sec:pretraining}).
We consider four domains (biomedical and computer science publications, news, and reviews; \sect{sec:dapt}) and eight classification tasks (two in each domain).
For targets that are not already in-domain for \roberta, our experiments show that continued pretraining on the domain  %\cite[
(which we refer to as {\em domain-adaptive pretraining} or \textbf{\dapt}) %, following][]{logeswaran-etal-2019-zero}
consistently improves performance on tasks from the target domain, in both high- and low-resource settings.  
%We use \roberta's masked language modeling loss on the domain corpus as a gauge of whether the corpus is out-of-domain for the model.  

%of the effect of target domain as well as task-specific data on adaptation of pretrained language models. We consider four domains (news, reviews, medical, computer science) and eight classification tasks (two in each domain). We use the masked language modeling loss a proxy to determine whether a given data is out-of-domain for \roberta. We then show that further training of the language model (henceforth, \textbf{domain-adaptive pretraining}) on such domains leads to performance gains.

% \nascomment{reworked here}
%that are essentially genres \swabha{Is the notion of a genre clear to a general reader or are we introducing terminology that might be more confusing?}\ana{+1 to avoid introducing genres.}\kyle{``genre'' is already used to define domain in the first intro paragraph},  
%\nascomment{I think this is a bit better}

% \swabha{not sure ``inductively'' is a common usage in this context} \nascomment{reworked, better now?} 

Above, we consider domains defined around genres and forums, but it is also possible to induce a domain from a given corpus used for a task, such as the one used in supervised training of a model.
This raises the question of whether pretraining on a corpus more directly tied to the \emph{task} can further improve performance.
We study how domain-adaptive pretraining compares to %what we term
{\em task-adaptive pretraining}, or \textbf{\tapt}, on a smaller but directly task-relevant corpus: the unlabeled task dataset (\sect{sec:tapt}), drawn from the \emph{task distribution}.  
Task-adaptive pretraining has been shown effective \cite{howard-ruder-2018-universal},
but is not typically used with the most recent models. 
%Despite the fact that task-adaptive pretraining is not current practice, %\swabha{This clause repeats what was said in the previous sentence}
We find that \tapt provides a large performance boost for \roberta, with or without domain-adaptive pretraining.
%new state-of-the-art results on two out of the five tasks where a comparable state of the art has been reported in the literature, and competitive performance on the remaining two \nascomment{three?} tasks. \nascomment{add a clause about the other three}\ana{done} \nascomment{numbers don't add up ... $2 + 2 \not = 5$.  maybe reword to say ``competitive performance on all five tasks, surpassing the state of the art on two,'' and clarify how our method is simpler/cheaper/something than the existing sota?} 

Finally, we show that the benefits from task-adaptive pretraining increase when we have additional unlabeled data %that (i) is drawn from the same distribution as the labeled data, and 
from the task distribution that has been \textit{manually curated}  
% \swabha{italicized this, since this is not commonly written}
by task designers or annotators. 
Inspired by this success, we propose ways to automatically select additional task-relevant unlabeled text, and show how this improves performance in certain low-resource cases (\sect{sec:more-tapt}).
On all tasks, our results using adaptive pretraining techniques are competitive with the state of the art.

%Inspired by the success of task-adaptive pretraining, even with small amounts of unlabeled data, we explore cases where there is an availability of more data from the task distribution, but not necessarily labeled.
%Such datasets are normally human-curated, and costly annotation efforts are exercised only on a small fraction of the data.
%We find large benefits from adapting to this oracle human-curated unlabeled corpora, available for some of the tasks.
%Finally, we explore some simple data selection strategies that work well in practice, and inch us closer towards this upper bound, without the need for an oracle dataset (\sect{sec:more-tapt}).

In summary, our contributions include: 
\begin{compactitem}
    \item a thorough analysis of 
    % the interplay between 
    domain- and task-adaptive pretraining across four domains and eight tasks, spanning low- and high-resource settings;
    \item an investigation into the transferability of adapted LMs across domains and tasks; and
    %\doug{need to expand what this means} of task-adaptation to other tasks in the same domain.
    \item  a study highlighting the importance of pretraining on human-curated datasets, and a simple data selection strategy to automatically approach this performance.
\end{compactitem}
% \swabha{I feel we could include the second point commented above, since the data selection part is as big a contribution as cross-\tapt stuff?}
Our code as well as pretrained models for multiple domains and tasks are publicly available.\footnote[1]{\smaller \url{https://github.com/allenai/dont-stop-pretraining}}

\begin{table*}[t!]
\centering
\small
% \resizebox{\textwidth}{!}{
\begin{tabular}{m{3cm}m{7cm}rrrr}
\toprule
\bf Domain & \bf Pretraining Corpus & \bf \# Tokens &\bf Size & \bf $\mathcal{L}_{\textsc{RoB.}}$ & \bf $\mathcal{L}_{\textsc{\dapt}}$\\
\midrule
\med      
& 2.68M  full-text papers from  \gorc \citep{lo2019gorc}  
& 7.55B 
& 47GB
& 1.32
& 0.99 \\
% \midrule[0.03em]
\cs 
& 2.22M full-text papers from \gorc \citep{lo2019gorc} 
& 8.10B  
& 48GB 
& 1.63
& 1.34 \\

% \midrule[0.03em]
\news
& 11.90M articles from \realnews \cite{zellers2019defending}%\footnote{similar to \ccnews sub-corpus used in \roberta pretraining}
& 6.66B  
& 39GB
& 1.08 
& 1.16 \\
% \midrule[0.03em]
\reviews
& 24.75M \amazon \citep{he2016ups} 
% \kyle{question, we didnt actually use all 143M, only trained up to 24.75M. so should we report actual used number?}  
& 2.11B 
& 11GB 
& 2.10
& 1.93 \\
\midrule[0.03em]
\roberta (baseline) & see Appendix \sect{sec:roberta_pretraining} 
& N/A 
& 160GB
& $^\ddagger$1.19
& - \\
\bottomrule
\end{tabular}
% }
\caption{List of the domain-specific unlabeled datasets. In columns 5 and 6, we report \roberta's masked LM loss on 50K randomly sampled held-out documents from each domain before ($\mathcal{L}_{\textsc{RoB.}}$) and after ($\mathcal{L}_{\textsc{\dapt}}$) \dapt (lower implies a better fit on the sample). $\ddagger$ indicates that the masked LM loss is estimated on data sampled from sources \textit{similar} to \roberta's pretraining corpus. 
% \swabha{Given that \roberta and others actually list the size in GBs, should we also do the same? } \nascomment{not a bad idea if straightforward}
%In Column 5, we report \roberta's masked LM loss on 50K randomly sampled held-out documents from each domain (lower implies a better fit). % \nascomment{check!}). 
%$\ddagger$ indicates that the masked LM loss is estimated on data sampled from sources \textit{similar} to \roberta's pretraining corpus. 
% \nascomment{instead of ``N/A'' for roberta \# tokens, maybe use \approx and give an estimate} \nascomment{commented out your footnote; you can't put footnotes in tables, they won't render without special tricks; I thought the fn was redundant; if you disagree move that point into caption}
}
\label{tab:domain_datasets}
\end{table*}

\section{Background: Pretraining}
\label{sec:pretraining}

Learning for most NLP research systems since 2018 consists of training in two stages.
First, a neural language model (LM), often with millions of parameters, is trained on large unlabeled corpora. 
The word (or wordpiece; \citealt{wu2016google}) representations
% parametric 
learned in the \textit{pretrained} model are then reused in supervised training for a downstream task, with optional updates (\textit{fine-tuning}) of the representations and network from the first stage.
% alongside any additional task-specific parameters.
% In this section we start a brief background on the design choices we adopt from \roberta (\sect{sec:roberta}).

%\subsection{\roberta} 
\label{sec:roberta}
% \nascomment{deleted subsection label since there are no additional subsections in this section}

One such pretrained LM is \roberta \cite{Liu2019RoBERTaAR}, which 
% , which is preceded by others such as \bert \cite{devlin-etal-2019-bert}, ELMo \cite{peters-etal-2018-deep}, and GPT \cite{radford2018improving}.
uses the same transformer-based architecture \cite{vaswani2017attention} as its predecessor, BERT \cite{devlin-etal-2019-bert}.
It is trained with a masked language modeling objective  (i.e., cross-entropy loss on predicting randomly masked tokens). %~\cite{devlin-etal-2019-bert}.
The unlabeled pretraining corpus for \roberta contains over 160 GB of uncompressed raw text from different English-language corpora (see Appendix \sect{sec:roberta_pretraining}). 
%As a result of the enormity of this corpus, as well as a careful choice of training strategies, \swabha{can cut this first part}
\roberta attains better performance on an assortment of tasks than its predecessors, making it our baseline of choice.
% Informed by all these factors, we select \roberta as our base model.

%The key assumption behind using a very large pretraining corpus is that, given an adequate amount of diverse data to train on, the representations from the LMs are powerful enough to generalize to a variety of tasks across domains. Our work questions the extent of this generalization.

Although \roberta's pretraining corpus is derived from multiple sources, it has not yet been established if these sources are diverse enough to generalize to most of the variation in the English language. 
In other words, we would like to understand what is out of \roberta's domain. 
Towards this end, we explore further adaptation by continued pretraining of this large LM into two categories of unlabeled data: (i) large corpora of domain-specific text (\sect{sec:dapt}), and (ii) available unlabeled data associated with a given task (\sect{sec:tapt}).

\section{\daptfull}
\label{sec:dapt}

%Although \roberta's pretraining corpus is derived from multiple sources, it has not yet been established if these sources are diverse enough to generalize to most of the variation in the English language. We would like to understand what is out-of-domain for \roberta. 

Our approach to domain-adaptive pretraining (\dapt) is straightforward---we continue pretraining \roberta on a large corpus of unlabeled domain-specific text. 
The four domains we focus on are biomedical (\med) papers, computer science (CS) papers, newstext from \realnews, and \amazon. 
We choose these domains because they have been popular in previous work, and datasets for text classification are available in each. 
Table \ref{tab:domain_datasets} lists the specifics of the unlabeled datasets in all four domains, as well as \roberta's training corpus.\footnote{For \med and \cs, we used an internal version of \textsc{S2ORC} that contains papers that cannot be released due to copyright restrictions.}

\subsection{Analyzing Domain Similarity}
\label{sec:domain_boundaries}

% {\color{red} [\textbf{Is the following paragraph true anymore given Reviews result??]}}
% We hypothesize that these domains are relatively distant for \roberta, with the exception of \realnews, which according to the authors, has a data distribution similar to the \ccnews corpus, used to train \roberta. 
%We include the news domain in order to see whether there are additional benefits of exposure to even more news data.

% \label{sec:domain_boundaries}

% Our analysis of \dapt is predicated on our intuitions about how corpora are situated under domains. 
% For example, to perform \dapt for \helpful, we only adapt to a large corpus of \amazon, and do not adapt to \news articles. 
% However, we note that such strict delineations of domain may not reflect the overlapping nature of these data distributions.

% \swabha{This seems like a big statement to be made without evidence...perhaps reference the improvements in the undapt settings to substantiate this?}

\begin{figure}[t]
 \hspace{-5mm}\includegraphics[scale=0.50]{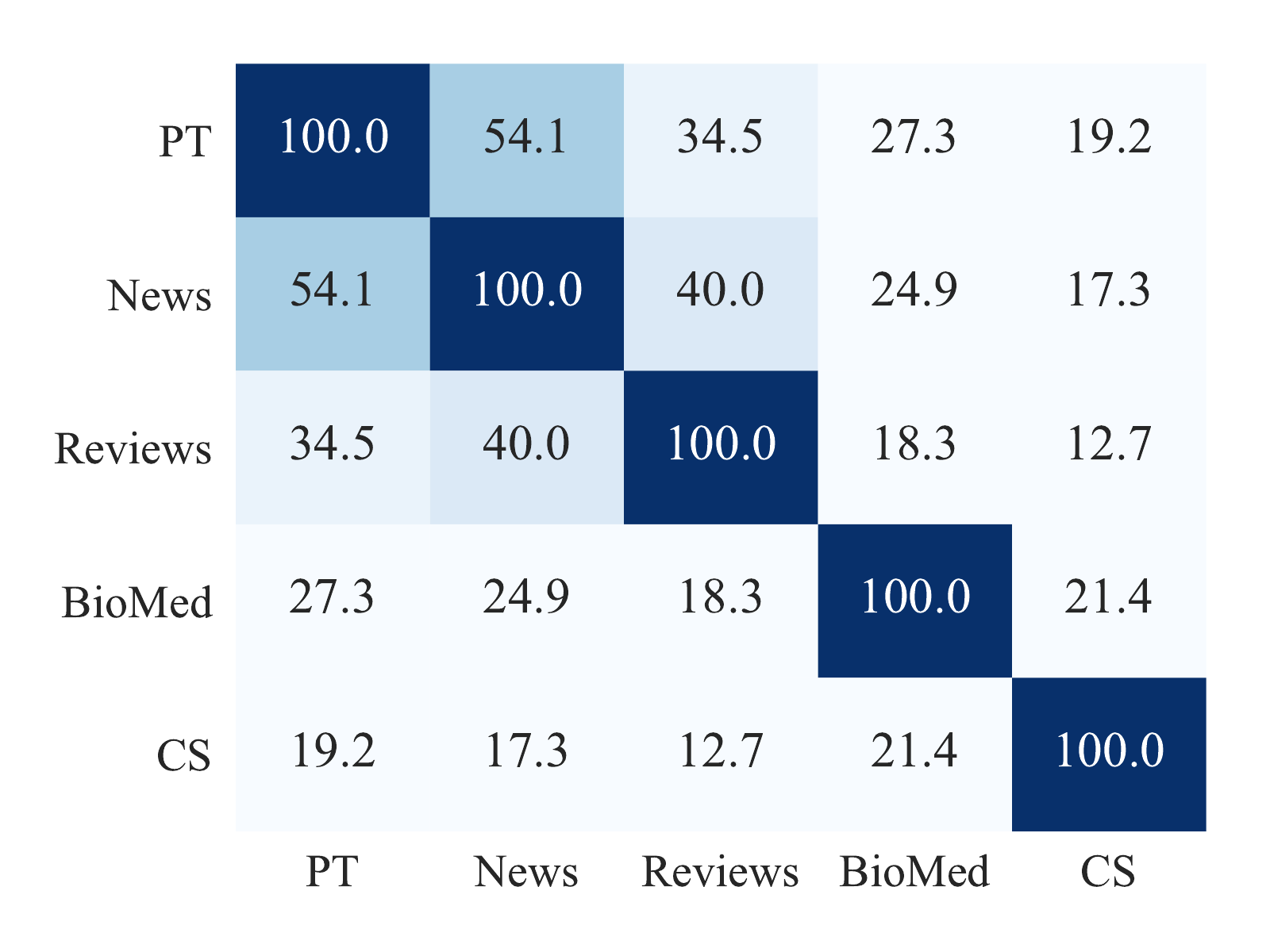}
 \caption{Vocabulary overlap (\%) between domains. \PT denotes a sample from sources similar to \roberta's pretraining corpus. 
 Vocabularies for each domain are created by considering the top 10K most frequent words (excluding stopwords) in documents sampled from each domain.} 
 \label{fig:vocabulary_overlap}
\end{figure}

Before performing \dapt, we attempt to quantify the similarity of the target domain to \roberta's pretraining domain.
We consider domain vocabularies containing the top 10K most frequent unigrams (excluding stopwords) in comparably sized random samples of held-out documents in each domain's corpus. 
% \swabha{Did you remove common stopwords? If not, standford-nlp (and maybe spacy) must offer lists for stopword removal. If we are considering the most frequent words, we must do this.}
We use 50K held-out documents for each domain other than \reviews, and 150K held-out documents in \reviews, since they are much shorter. 
We also sample 50K documents from sources similar to \roberta's pretraining corpus (i.e., \books, \textsc{Stories}, \wiki, and \realnews) to construct the pretraining domain vocabulary, since the original pretraining corpus is not released. 
Figure \ref{fig:vocabulary_overlap} shows the vocabulary overlap across these samples. 
We observe that \roberta's pretraining domain has strong vocabulary overlap with \news and \reviews, while \cs and \med are far more dissimilar to the other domains.
This simple analysis suggests the degree of benefit to be expected by adaptation of \roberta to different domains---the more dissimilar the domain, the higher the potential for \dapt.

\subsection{Experiments}
\label{sec:dapt_experiments}

Our LM adaptation follows the settings prescribed for training \roberta. %in \citet{Liu2019RoBERTaAR}.
We train \roberta on each domain for 12.5K steps, which amounts to single pass on each domain dataset, on a v3-8 TPU; see other details in Appendix \ref{sec:training}.
This second phase of pretraining results in four domain-adapted LMs, one for each domain. We present the masked LM loss of \roberta on each domain before and after \dapt in Table \ref{tab:domain_datasets}. We observe that masked LM loss decreases in all domains except \news after \dapt, where we observe a  marginal increase. We discuss  cross-domain masked LM loss in Appendix \sect{sec:mlm_study_appendix}.

Under each domain, we consider two text classification tasks, as shown in Table \ref{tab:data_tasks}.
Our tasks represent both high- and low-resource ($\le$ 5K labeled training examples, and no additional unlabeled data) settings. 
For \hp, we use the data splits from \citet{Beltagy2020Longformer}. 
For \rct, we represent all sentences in one long sequence for simultaneous prediction.

\begin{table*}[t]
\centering
\small
% \resizebox{\textwidth}{!}{
\begin{tabular}{lllrrrrr}
\toprule
\textbf{Domain}
& \textbf{Task}            
& \textbf{Label Type} 
& \textbf{Train (Lab.)} 
& \textbf{Train (Unl.)} 
& \textbf{Dev.} 
& \textbf{Test} 
& \textbf{Classes}      \\
\midrule
\multirow{2}{*}{\med} 
& \chemprot 
& relation classification
& 4169  
& -      
& 2427  
& 3469 
&       13    \\
& $^\dagger$\rct            
& abstract sent. roles 
& 18040    
& - 
& 30212 
& 30135 
& 5 \\
\midrule[0.03em]
\multirow{2}{*}{\cs}
& \arccite  
& citation intent 
& 1688    
& -        
&  114 
& 139 
& 6                     \\
& \sciie     
&  relation classification     
& 3219 
& -  
& 455  
& 974        
& 7    \\
\midrule[0.03em]
\multirow{2}{*}{\news}
& \hp 
& partisanship 
& 515 
& 5000   
& 65
& 65 
& 2             \\
& $^\dagger$\ag       
& topic  
& 115000    
& -   
& 5000   
& 7600  
& 4      \\
\midrule[0.03em]
\multirow{2}{*}{\reviews} 
& $^\dagger$\helpful
& review helpfulness 
& 115251 
& - 
& 5000 
& 25000  
& 2    \\
& $^\dagger$\imdb     
& review sentiment  
& 20000 
& 50000  
& 5000 
& 25000 
& 2       \\ 
\bottomrule
\end{tabular}
% }
\caption{Specifications of the various target task datasets.
$\dagger$ indicates high-resource settings.
Sources: \chemprot \cite{Kringelum2016ChemProt30AG}, \rct \cite{Dernoncourt2017PubMed2R}, \arccite \cite{Jurgens2018MeasuringTE}, \sciie \cite{Luan2018MultiTaskIO}, \hp \cite{stein:2019h}, \ag \cite{Zhang2015CharacterlevelCN}, \helpful \cite{mcauley2015image}, \imdb \cite{Maas2011LearningWV}. 
}
\label{tab:data_tasks}
\end{table*}

\begin{table}[t]
\centering
% \resizebox{\columnwidth}{!}{
\begin{tabular}{p{0.7cm}p{1.5cm}rrr}
\toprule
%  & &  \multicolumn{2}{c}{Additional Pretraining}\\
% \cline{3-4}
\textbf{Dom.} &   
\textbf{Task} 
& {\bf \textsc{RoBa.}} & \textbf{\dapt} & \bf $\lnot$\dapt\\
\midrule 
\multirow{2}{*}{\small \textsc{BM}} &   
\chemprot 
& 81.9$_{1.0}$ & \bf 84.2$_{0.2}$  & 79.4$_{1.3}$ \\
&
$^\dagger$\rct  
& 87.2$_{0.1}$ & \bf 87.6$_{0.1}$  & 86.9$_{0.1}$ \\ 
\midrule[0.03em]
\multirow{2}{*}{\small \cs} & 
{\footnotesize \arccite}
& 63.0$_{5.8}$ & \bf 75.4$_{2.5}$  & 66.4$_{4.1}$ \\
& 
\sciie  
& 77.3$_{1.9}$ & \bf 80.8$_{1.5}$  &  79.2$_{0.9}$ \\
\midrule[0.03em]
\multirow{2}{*}{\small \news}& 
\hpshort 
& 86.6$_{0.9}$ &   \bf 88.2$_{5.9}$  & 76.4$_{4.9}$ \\
& 
$^\dagger$\ag
& \bf 93.9$_{0.2}$ & \bf 93.9$_{0.2}$  & 93.5$_{0.2}$ \\
\midrule[0.03em]
% \multirow{2}{*}
\multirow{2}{*}{\textsc{Rev.}} & 
$^\dagger$\textsc{Helpful}.
& 65.1$_{3.4}$ & \bf 66.5$_{1.4}$  &  65.1$_{2.8}$ \\
& 
$^\dagger$\imdb 
& 95.0$_{0.2}$ & \bf 95.4$_{0.2}$   &  94.1$_{0.4}$ \\
\bottomrule
\end{tabular}
% }
\caption{Comparison of \roberta (\textsc{RoBa.}) and \dapt to adaptation to an \emph{irrelevant} domain ($\lnot$ \dapt). Reported results are test macro-$F_1$, except for \chemprot and \rct, for which we report micro-$F_1$, following \citet{Beltagy2019SciBERTAP}. We report averages across five random seeds, with standard deviations as subscripts.
$\dagger$ indicates high-resource settings. Best task performance is boldfaced. %For \news, we chose \cs as the irrelevant domain, for \reviews, we chose \med as the irrelevant domain, and for \cs and \med, we chose \news as the irrelevant domain. 
See \sect{sec:undomain} for our choice of irrelevant domains. 
% \swabha{It would be nice to add the undomain domain to each row here...but it's a \latex nightmare to confine it to a single column.}
}
\label{tab:undomain}
\end{table}

\paragraph{Baseline}
As our baseline, we use an off-the-shelf \roberta-base model and perform supervised fine-tuning of its parameters for each classification task. 
%For \hp, the best reported performance  \cite{Beltagy2020Longformer} uses a modified Transformer language model for long documents.
% or on \rct\todo{Come back to RCT}\footnote{With training set with $\sim$180K examples, the best reported performance is 92.6 \cite{jin-szolovits-2018-hierarchical}, and $92.0$ if sentences are classified without taking the surrounding sentences into the account (our setting). }
% We downsampled the training data to only 500 examples.\todo{Go back to RCT.}} 
On average, \roberta is not drastically behind the state of the art (details in Appendix \sect{sec:sota}), and serves as a good baseline since it provides a single LM to adapt to different domains. 

% \paragraph{Masked LM Loss}  \nascomment{propose moving this to the end of 3.2 and reframe as analysis; bigger improvements to MLM are (I suspect) predictive of bigger gains from DAPT}
% A more expensive method for analyzing domain similarity entails computing the masked LM loss of \roberta on the sample texts from each domain before ($\mathcal{L}_\roberta$) and after ($\mathcal{L}_\dapt$) \dapt. We calculate the loss on the same held-out domain samples from \sect{sec:domain_boundaries}.
% Results in Table \ref{tab:masked_lm_study_main_deltaL} 
% % \suchin{why is this showing up as Table 3.2 and not Table 3???} 
% align with the intuition that the decrease in loss after \dapt is expected to be smaller for domains closer to the ones in \roberta's training corpus. 
% Indeed, we see that \news has the least amount of domain novelty, whereas \med, \cs, and \reviews are more distant. 
% On the other hand, $\mathcal{L}_\dapt$ measures how well the language model fits a target domain after \dapt. We observe that $\mathcal{L}_\dapt$ decreases in all domains except \news, where we observe a marginal increase. 
% We discuss cross-domain masked LM loss in Appendix \sect{sec:mlm_study_appendix}.

\paragraph{Classification Architecture} Following standard practice \citep{devlin-etal-2019-bert} we pass the final layer \texttt{[CLS]} token representation to a task-specific feedforward layer for prediction (see Table \ref{tab:roberta_text_classification_hyperparameters} in Appendix for more hyperparameter details).

\paragraph{Results} 
% \swabha{I think these results need to be presented after the discussion of domain similarity, since we talk about more and less distant domains.}

 Test results are shown under the \dapt column of Table \ref{tab:undomain} (see Appendix \sect{sec:dev_set_results} for validation results). 
We observe that \dapt improves over \roberta in all domains. 
For \med, \cs, and \reviews, we see consistent improvements over \roberta, demonstrating the benefit of \dapt when the target domain is more distant from \roberta's source domain. 
The pattern is consistent across high- and low- resource settings. %Consistent with our domain shift findings in \sect{sec:domain_shift}, 
Although \dapt does not increase performance on \ag, the benefit we observe in \hp suggests that \dapt may be useful even for tasks that align more closely with \roberta's source domain.

\begin{table*}[t]
\centering
\small
\resizebox{\textwidth}{!}{
\begin{tabular}{>{\raggedright}p{11cm}p{6cm}}
\toprule 
\textbf{\imdb review}  & \textbf{\realnews article}\\
\midrule 
\colorbox{orange!30!}{“The Shop Around the Corner“} is one of the \colorbox{orange!30!}{great films} from director \colorbox{orange!30!}{Ernst Lubitsch}. In addition to the talents of \colorbox{orange!30!}{James Stewart} and \colorbox{orange!30!}{Margaret Sullavan}, it's filled with a terrific cast of top character actors such as Frank Morgan and Felix Bressart. [...] The makers of \colorbox{orange!30!}{“You've Got Mail“} claim their film to be a \colorbox{orange!30!}{remake}, but that's just nothing but a lot of inflated self praise. Anyway, if you have an affection for romantic \colorbox{orange!30!}{comedies} of the \colorbox{orange!30!}{1940}'s, you'll find \colorbox{orange!30!}{“The Shop Around the Corner“} to be nothing short of wonderful. Just as good with repeat viewings. & [...] Three great festive films... The Shop Around the Corner (1940) Delightful Comedy by Ernst Lubitsch stars James Stewart and Margaret Sullavan falling in love at Christmas. Remade as You’ve Got Mail. [...] \\
\midrule
%\multicolumn{2}{>{\raggedright}p{16.5cm}}{\textbf{\imdb review with a partisan language}}\\
%\midrule 
%\multicolumn{2}{>{\raggedright}p{16.5cm}}{A film like Amazing Grace and Chuck is a perfect example of how \colorbox{orange!30!}{the left in this country just doesn't get it}. They never did. \colorbox{orange!30!}{And liberalism continues to slip further and further into political oblivion}. This film deals with a little league baseball star who decides to stop playing ball as a protest to the existence of nuclear weapons. [...] Neither side during the cold war was crazy enough to fire a single missile. Without these weapons, who knows what might have happened between the USA and USSR. The makers of this film obviously intended for kids in America to take up their cause and follow in Chuck's footsteps. \colorbox{orange!30!}{Kids in America however are more intelligent than the left wing kooks who created this dreck}. [...]}\\
\textbf{\helpful review}  & \textbf{\realnews article}\\
\midrule
Simply the Best! I've owned countless Droids and iPhones, but this one destroys them all. \colorbox{orange!30!}{Samsung} really nailed it with this one, extremely \colorbox{orange!30!}{fast}, very pocketable, gorgeous \colorbox{orange!30!}{display}, exceptional \colorbox{orange!30!}{battery life}, good audio quality, perfect GPS \& WiFi performance, transparent status bar, \colorbox{orange!30!}{battery} percentage, ability to turn off soft key lights, superb \colorbox{orange!30!}{camera} for a smartphone and more! [...]
% Forget about the new iPhone or any other Droid for that matter, this one is absolute perfection! 
&  
We’re living in a world with a new Samsung. [...] more on battery life later [...] Exposure is usually spot on and focusing is very fast. [...] The design, display, camera and performance are all best in class, and the phone feels smaller than it looks. [...]\\
\bottomrule
\end{tabular}}
\caption{Examples that illustrate how some domains might have overlaps with others, leading to unexpected positive transfer. 
We highlight expressions in the reviews that are also found in the \realnews articles.
%These examples help us understand why news-\dapt helps tasks in the Reviews domain.
%and a clearly partisan language in the second review.
} 
\label{tab:analysis_realnews_news}
\end{table*}

%\subsection{$\lnot$\dapt Baseline}
\subsection{Domain Relevance for \dapt}
\label{sec:undomain}

Additionally, we compare \dapt against a setting where for each task, we adapt the LM to a domain \textbf{outside} the domain of interest.
This controls for the case in which the improvements over \roberta might be attributed simply to exposure to more data, regardless of the domain.
In this setting, for \news, we use a \cs LM; for \reviews, a \med LM; for \cs, a \news LM;  for \med, a \reviews LM. We use the vocabulary overlap statistics in Figure \ref{fig:vocabulary_overlap} to guide these choices.

% {\color{red} [\textbf{highlight DAPT is much better than not DAPT, but there are some cases where training on \emph{any} additional data can still be more useful than doing nothing.]}}
Our results are shown in Table \ref{tab:undomain}, where the last column ($\lnot$\dapt) corresponds to this setting. 
For each task, \dapt significantly outperforms adapting to an irrelevant domain, suggesting the importance of pretraining on domain-relevant data. 
Furthermore, we generally observe that $\lnot$\dapt results in worse performance than even \roberta on end-tasks. Taken together, these results indicate that in most settings, exposure to more data without considering domain relevance is detrimental to end-task performance. However, there are two tasks (\sciie and \arccite) in which $\lnot$\dapt marginally \emph{improves} performance over \roberta. This may suggest that in some cases, continued pretraining on any additional data is useful, as noted in \citet{baevski2019cloze}.

\subsection{Domain Overlap}

Our analysis of \dapt is based on prior intuitions about how task data is assigned to specific domains. For instance, to perform \dapt for \helpful, we only adapt to \amazon, but not to any \realnews articles. However, the gradations in Figure \ref{fig:vocabulary_overlap} suggest that the boundaries between domains are in some sense fuzzy; for example, 40\% of unigrams are shared between \reviews and \news.
As further indication of this overlap, we also qualitatively identify documents that overlap cross-domain: in Table \ref{tab:analysis_realnews_news}, we showcase reviews and \realnews articles that are similar to these reviews (other examples can be found in Appendix \sect{sec:appendix_domain_transfer}). In fact, we find that adapting \roberta to \news not as harmful to its performance on \reviews tasks (\dapt on \news achieves 65.5$_{2.3}$ on \helpful and 95.0$_{0.1}$ on \imdb). 

% \nascomment{reworked here, please check} 
Although this analysis is by no means comprehensive, it indicates that the factors that give rise to observable domain differences are likely not mutually exclusive. It is possible that pretraining beyond conventional domain boundaries could result in more effective \dapt; we leave this investigation to future work. In general, the provenance of data, including the processes by which corpora are curated, must be kept in mind when designing pretraining procedures and creating new benchmarks that test out-of-domain generalization abilities.

\begin{table*}[t]
\centering
% \resizebox{0.9\textwidth}{!}{
\begin{tabular}{llcccc}
\toprule
 &  & & \multicolumn{3}{c}{Additional Pretraining Phases}\\
\cline{4-6}
\textbf{Domain} &   \textbf{Task} 
& \bf \roberta & \textbf{\dapt} & \textbf{\tapt} & \textbf{\dapt + \tapt} \\ %& \boldmath{\lnot}\textbf{\dapt}\\
\midrule 
\multirow{2}{*}{\med} &   \chemprot 
& 81.9$_{1.0}$& 84.2$_{0.2}$& 82.6$_{0.4}$& \bf 84.4$_{0.4}$\\ %& 80.6$_{1.2}$\\
& $^\dagger$\rct  
& 87.2$_{0.1}$ & 87.6$_{0.1}$ & 87.7$_{0.1}$ & \bf 87.8$_{0.1}$ \\ %& 72.7$_{0.7}$\\ 
\midrule[0.03em]
\multirow{2}{*}{\cs}& \arccite 
& 63.0$_{5.8}$& 75.4$_{2.5}$ & 67.4$_{1.8}$&  \bf 75.6$_{3.8}$\\ %& 67.7$_{5.7}$\\
& \sciie
& 77.3$_{1.9}$& 80.8$_{1.5}$& 79.3$_{1.5}$ & \bf 81.3$_{1.8}$\\ %& 80.0$_{3.0}$\\
\midrule[0.03em]
\multirow{2}{*}{\news}& \hp 
% 82.1
& 86.6$_{0.9}$& 88.2$_{5.9}$& \bf 90.4$_{5.2}$ & 90.0$_{6.6}$\\ %& 79.0$_{4.1}$\\
& $^\dagger$\ag 
& 93.9$_{0.2}$& 93.9$_{0.2}$& 94.5$_{0.1}$& \bf 94.6$_{0.1}$\\ %& 93.4$_{0.2}$\\
\midrule[0.03em]
\multirow{2}{*}{\reviews} & $^\dagger$\helpful
& 65.1$_{3.4}$& 66.5$_{1.4}$& 68.5$_{1.9}$&  \bf 68.7$_{1.8}$\\ %&  66.9$_{3.2}$\\
& $^\dagger$\imdb 
& 95.0$_{0.2}$& 95.4$_{0.1}$ & 95.5$_{0.1}$ & \bf 95.6$_{0.1}$\\ %&  94.5$_{0.3}$\\
\bottomrule
\end{tabular}
% }
\caption{Results on different phases of adaptive pretraining compared to the baseline \roberta (col.~1).
Our approaches are \dapt (col.~2, \sect{sec:dapt}), \tapt (col.~3, \sect{sec:tapt}), and a combination of both (col.~4). Reported results follow the same format as Table \ref{tab:undomain}. State-of-the-art results we can compare to: \chemprot (84.6), \rct (92.9), \arccite (71.0), \sciie (81.8), \hp (94.8), \ag (95.5), \imdb (96.2); references in \sect{sec:sota}. 
% \nascomment{is the boldface misplaced in the last row?}
}
\label{tab:main}
\end{table*}

\section{\taptfull}
\label{sec:tapt}

% \nascomment{reworked}
% When a dataset is constructed for training and 
% testing a specific NLP capability, it necessarily covers only a small subset of the text available within the broader domain.
% Often, the training and test corpora for a task are carefully curated to capture a niche of language that is rich in a phenomena of interest.
% For example, the \chemprot task \cite{Krallinger2017OverviewOT}  is aimed at automatically extracting relationships between chemical compounds and proteins. 
% Its training and test corpora consist of abstracts from chemistry-related papers---a small fraction of the broader domain of all medical text.  
% In fact, the selection criteria for the \textsc{chemdner} corpus \cite{krallinger2015chemdner} used in \chemprot restricts the data in a variety of specific ways: it includes only abstracts of articles from hand-selected categories of PubMed papers with high impact factor, published in recent years.  

% In such cases where the task data is a narrowly-defined subset of the domain, pretraining on it directly or on similar task-relevant data may be helpful.
Datasets curated to capture specific tasks of interest tend to cover only a subset of the text available within the broader domain.  
For example, the \chemprot dataset for extracting relations between chemicals and proteins focuses on abstracts of recently-published, high-impact articles from hand-selected PubMed categories \cite{Krallinger2017OverviewOT,krallinger2015chemdner}. We hypothesize that such cases where the task data is a narrowly-defined subset of the broader domain, pretraining on the task dataset itself or data relevant to the task may be helpful.
% } \nascomment{why is this paragraph orange?}

% \kyle{Consider changing ``necessarily cover'' to ``often cover''.  It's not obvious from the Chemprot example that the dataset \textit{needed} to be restricted in that fashion, only that it was.}
% \doug{Not sure who wanted 'necessarily' there, I don't think it was me but I can see the argument for it.  Nonetheless on balance I agree with Kyle.  Changed it to 'tend to' for now.  Rewrite was good, I made minor tweaks b/c it's only CHEMDNER that's restricted in those ways, and the paper about that data set makes kind of a big deal about that curation process; ChemProt just uses the CHEMDNER data in addition to other (differently restricted, but in less well-defined way) data.  I decided to not mention CHEMDNER but just cite that paper in addition the ChemProt at the end of the sentence, if a reader actually cares about this they can sort out the details on their own.}

{\em Task-adaptive pretraining} (\tapt) refers to pretraining on the unlabeled training set for a given task; prior work has shown its effectiveness \cite[e.g.][]{howard-ruder-2018-universal}.  
Compared to domain-adaptive pretraining (\dapt; \S\ref{sec:dapt}), the task-adaptive approach strikes a different trade-off: it uses a far smaller pretraining corpus, but one that is much more task-relevant (under the assumption that the training set represents aspects of the task well).  
This makes \tapt much less expensive to run than \dapt , and as we show in our experiments, the performance of \tapt is often competitive with that of \dapt.  
\begin{table*}[t]
\centering
\resizebox{\textwidth}{!}{  
\begin{tabular}{l|ll}
\toprule
% \multirow{3}{*}{\rotatebox{90}{\textsc{BioMed}}}   
\med  
& \rct                             
&  \chemprot  \\
%\cmidrule{2-4} 
\midrule
\tapt      
& 87.7$_{0.1}$     
& 82.6$_{0.5}$ \\
Transfer-\tapt 
& 87.1$_{0.4}$ ($\downarrow$0.6)
& 80.4$_{0.6}$ ($\downarrow$2.2)  \\
\bottomrule
\toprule
%  \multirow{3}{*}{\rotatebox{90}{\textsc{ReNews}}} 
\news        
 & \hp                     
 &  \ag   \\    
%\cmidrule{2-4} 
\midrule
\tapt
& 89.9$_{9.5}$ 
& 94.5$_{0.1}$ \\
Transfer-\tapt
& 82.2$_{7.7}$ ($\downarrow$7.7) 
& 93.9$_{0.2}$ ($\downarrow$0.6)                    \\
\bottomrule
\end{tabular}
\quad
\begin{tabular}{l|ll}
\toprule
%  \multirow{3}{*}{\rotatebox{90}{\textsc{CompSc}}} 
\cs   
 &  \arccite                   
 & \sciie \\
%\cmidrule{2-4}  
\midrule
\tapt
& 67.4$_{1.8}$                
& 79.3$_{1.5}$ \\
Transfer-\tapt 
& 64.1$_{2.7}$ ($\downarrow$3.3) 
& 79.1$_{2.5}$ ($\downarrow$0.2) \\
\bottomrule
\toprule
% \multirow{3}{*}{\rotatebox{90}{\textsc{reviews}}} 
\reviews    
&  \helpful                         
& \imdb     \\
%\cmidrule{2-4} 
\midrule
\tapt    
& 68.5$_{1.9}$ 
& 95.7$_{0.1}$\\
 Transfer-\tapt 
& 65.0$_{2.6}$ ($\downarrow$3.5)    
& 95.0$_{0.1}$ ($\downarrow$0.7) \\
\bottomrule
\end{tabular}}

\caption{
% \nascomment{reworked to be more clear that this is not a negative result, and to tie back to the notion of task distribution introduced earlier} 
Though \tapt is effective (Table~\ref{tab:main}), it is harmful when applied \emph{across} tasks.  These findings illustrate differences in task distributions within a domain.}
\label{tab:transfer_test_results}
\end{table*}
\subsection{Experiments}
\label{sec:tapt-results}

Similar to \dapt, task-adaptive pretraining consists of a second phase of pretraining \roberta, but only on the available task-specific training data.
In contrast to \dapt, which we train for $12.5$K steps, we perform \tapt for 100 epochs. 
We artificially augment each dataset by randomly masking different words (using the masking probability of 0.15) across epochs. As in our \dapt experiments, we pass the final layer \texttt{[CLS]} token representation to a task-specific feedforward layer for classification  (see Table \ref{tab:roberta_text_classification_hyperparameters} in Appendix for more hyperparameter details).
% For \tapt, we examine the case in which the unlabeled task data comprises only training data used for supervised fine-tuning, which we call \emph{micro-\tapt}. 

Our results are shown in the \tapt column of Table~\ref{tab:main}. 
\tapt consistently improves the \roberta baseline for all tasks across domains.
Even on the news domain, which was part of \roberta pretraining corpus, \tapt improves over \roberta, showcasing the advantage of task adaptation.
Particularly remarkable are the relative differences between \tapt and \dapt.
\dapt is more resource intensive (see Table \ref{tab:speed} in \sect{sec:speed}), but \tapt manages to match its performance in some of the tasks, such as \sciie.
% which is much more resource-intensive, and \nascomment{something broken here!  I think you're trying to say that \dapt is more resource intensive, but \tapt manages to come close or match it in some cases; that is not what the current text says} yet seems to be almost as effective 
In \rct, \hp, \ag, \helpful, and \imdb, the results even exceed those of \dapt, highlighting the efficacy of this cheaper adaptation technique.

%\nascomment{something important that's not coming through:  how much cheaper \tapt is than \dapt.  
%yes  I can look at the numbers in the tables and guess, but it would be nice if you had either a column in the table giving a sense of the \tapt:\dapt ratio (in amount of data) or a statement of the range of that ratio, in the text.  
%also, in the text, translate that into real terms, e.g., ``In our experiments, the \tapt phase took between 5 and 10\% as much wall time as the \dapt phase.''}\ana{Added a pointer to Table 9 that comes later.}

\paragraph{Combined \dapt and \tapt}
\label{sec:dapt-tapt-results}

We investigate the effect of using both adaptation techniques together.
We begin with \roberta and apply \dapt then \tapt under this setting.
The three phases of pretraining add up to make this the most computationally expensive of all our settings (see Table \ref{tab:speed}). 
As expected, combined  domain- and task-adaptive pretraining achieves the best performance on all tasks (Table \ref{tab:main}).\footnote{Results on \hp match those of \tapt, within a standard deviation arising from the five seeds.}
% Moreover, on \chemprot, \arccite, \sciie, we obtain state-of-the-art performance. For \imdb and \ag, domain- and task-adaptive pretraining is competitive with existing state-of-the-art methods.  While we outperform all reported methods on \hp, we cannot compare to existing methods because the test set of this shared task is not publicly available. Finally, we establish a new benchmark for \helpful, on which no models have been evaluated before.
%\todo{Include performance reported in other papers to substantiate this claim.} \nascomment{this!}\swabha{done, please see table.}

Overall, our results show that \dapt followed by \tapt achieves the best of both worlds of domain and task awareness, yielding the best performance. While we speculate that \tapt followed by \dapt would be susceptible to catastrophic forgetting of the task-relevant corpus \cite{Yogatama2019LearningAE}, alternate methods of combining the procedures may result in better downstream performance. Future work may explore pretraining with a more sophisticated curriculum of domain and task distributions. 

% We speculate that \tapt followed by \dapt would be susceptible to catastrophic forgetting of the task-relevant corpus \cite{Yogatama2019LearningAE}.

% We considered alternate methods of combining \dapt and \tapt, such as interleaving the procedures. 
% \swabha{Hmm, not sure what ``considered'' means here---just a thought experiment or did we ever run these? I can't remember ever seeing these experiments, correct me if I'm wrong.}
% However, because the domain-relevant corpus is much larger than the task-relevant corpus, we surmise that the task-relevant signal is likely to be drowned out when interleaving the corpora. %during pretraining.
% \nascomment{reworked here, please check} 
% Furthermore, we speculate that \tapt followed by \dapt would be susceptible to catastrophic forgetting of the task-relevant corpus \cite{Yogatama2019LearningAE}. Thus, we recommend the \dapt, then \tapt sequence of pretraining. 
% Future work may explore pretraining with a more sophisticated curriculum of domain and task distributions.

% \nascomment{in case I didn't already note this:  need to clarify where those s.d.s are coming from, in the text}

\paragraph{Cross-Task Transfer}
\label{sec:ct_transfer}

% \doug{need to say why we're doing this, it comes a bit out of left field.}\ana{Added a clause.} 
We complete the comparison between \dapt and \tapt by exploring whether adapting to one task transfers to other tasks in the same domain. 
For instance, we further pretrain the LM using the \rct unlabeled data, fine-tune it with the \chemprot labeled data, and observe the effect.
We refer to this setting as Transfer-\tapt. 
%\nascomment{not clear what task you're using for \tapt and what task you're using for supervised training.  I think you're basically taking the other task within each domain, but don't make me guess that}\ana{I added an example.}
Our results for tasks in all four domains are shown in Table \ref{tab:transfer_test_results}.
We see that \tapt optimizes for single task performance, to the detriment of cross-task transfer.  
These results demonstrate that data distributions of tasks within a given domain might  differ. 
Further, this could also explain why adapting only to a broad domain is not sufficient, and why \tapt after \dapt is effective.

%\nascomment{bring this back to the notion of task distributions!!}\ana{There is my attempt. }\swabha{Also see end of this section.}

%In light of the previous section (\sect{sec:cd_transfer}), we investigate how the adaptation to \imdb affects the \hp task. The LM that is task-adapted to \imdb achieves the $F1$ score of $85.7_{2.3}$ on the \hp dataset (not shown in the table), outperforming the LM that is task-adapted to \hp which gets $83.9_{2.6}$. Therefore, although \tapt usually does not transfer well across tasks, there is a task-pair where it results in positive transfer. However, identifying this task pair required a good understanding of underlying data distributions.  \ana{This is my takeaway from this result, feel free to change.}

% \ana{Suggestion: make cross-transfer a paragraph of Sec 4.1 (Experiments), and add a new paragraph after cross-task transfer saying “DAPT, TAPT, or both?” and move this disucssion there?}

\section{Augmenting Training Data for Task-Adaptive Pretraining}
\label{sec:more-tapt}

In \sect{sec:tapt}, we continued pretraining the LM for task adaptation using only the training data for a supervised task. 
Inspired by the success of \tapt, we next investigate another setting where a larger pool of unlabeled data from the task distribution exists, typically curated by humans. 

We explore two scenarios.
First, for three tasks (\rct, \hp, and \imdb) we use this larger pool of unlabeled data from an available human-curated corpus (\sect{sec:oracle}).
Next, we explore \emph{retrieving} related unlabeled data for \tapt, from a large unlabeled in-domain corpus, for tasks where extra human-curated data is unavailable (\sect{sec:selection}). 
\begin{table}[t]
\centering
\resizebox{\columnwidth}{!}{
% \scalebox{0.8}{
\begin{tabular}{ccp{1cm}p{1.5cm}}
\toprule
\multirow{2}{*}{Pretraining} &\textsc{BioMed} & \news & \reviews \\
& \rct-500        & \hpshort    & \imdb$^\dagger$         \\
\midrule
%-            & -            & 71.5$_{1.1}$ & 82.9$_{1.6}$ & 92.3$_{0.5}$ \\
% \dapt       & -            & $74.9_{0.6}$ & $82.7_{2.2}$ & $93.2_{0.5}$ \\
\tapt     & 79.8$_{1.4}$ & 90.4$_{5.2}$ & 95.5$_{0.1}$ \\
\dapt + \tapt & 83.0$_{0.3}$ & 90.0$_{6.6}$ & 95.6$_{0.1}$ \\
\midrule[0.03em]
\oracle & 83.4$_{0.3}$ & 89.9$_{9.5}$ & 95.7$_{0.1}$ \\
\dapt + \oracle & \bf{83.8$_{0.5}$} & \bf{92.1$_{3.6}$} & \bf{95.8$_{0.1}$} \\
\bottomrule
\end{tabular}
}
% }
\caption{Mean test set macro-$F_1$ (for \hpshort and \imdb) and micro-$F_1$ (for \rct-500), with \oracle across five random seeds, with standard deviations as subscripts.
$\dagger$ indicates high-resource settings. }
\label{tab:test_micro_vs_macro_task_adaptive_results}
\end{table}

\subsection{Human \oracle}
\label{sec:oracle}
Dataset creation often involves collection of a large unlabeled corpus from known sources.
This corpus is then downsampled to collect annotations, based on the annotation budget.
The larger unlabeled corpus is thus expected to have a similar distribution to the task's training data.
Moreover, it is usually available.
We explore the role of such corpora in task-adaptive pretraining.

\paragraph{Data}
We simulate a low-resource setting \rct-500, by downsampling the training data of the \rct dataset to 500 examples (out of 180K available), and treat the rest of the training data as unlabeled.
The \hp shared task \cite{stein:2019h} has two tracks: low- and high-resource. 
We use 5K documents from the high-resource setting as \oracle unlabeled data and the original low-resource training documents for task fine-tuning. 
For \imdb, we use the extra unlabeled data manually curated by task annotators, drawn from the same distribution as the labeled data \cite{Maas2011LearningWV}.

\paragraph{Results}  We compare \oracle to \tapt and \dapt+ \tapt in Table \ref{tab:test_micro_vs_macro_task_adaptive_results}.
\oracle further improves our prior results from \sect{sec:tapt} across all three datasets.
Applying \oracle after adapting to the domain results in the largest boost in performance on all tasks; in \hp, \dapt + Curated-\tapt is within standard deviation of \oracle.  
% \nascomment{note to self - check here}
Moreover, curated-\tapt achieves 95\% of the performance of \dapt+ \tapt with the fully labeled \rct corpus (Table \ref{tab:main}) with only 0.3\% of the labeled data. 
These results suggest that curating large amounts of data from the task distribution is extremely beneficial to end-task performance. 
We recommend that task designers release a large pool of unlabeled task data for their tasks to aid model adaptation through pretraining.

\begin{figure}[t]
     \centering
    \includegraphics[width=\columnwidth]{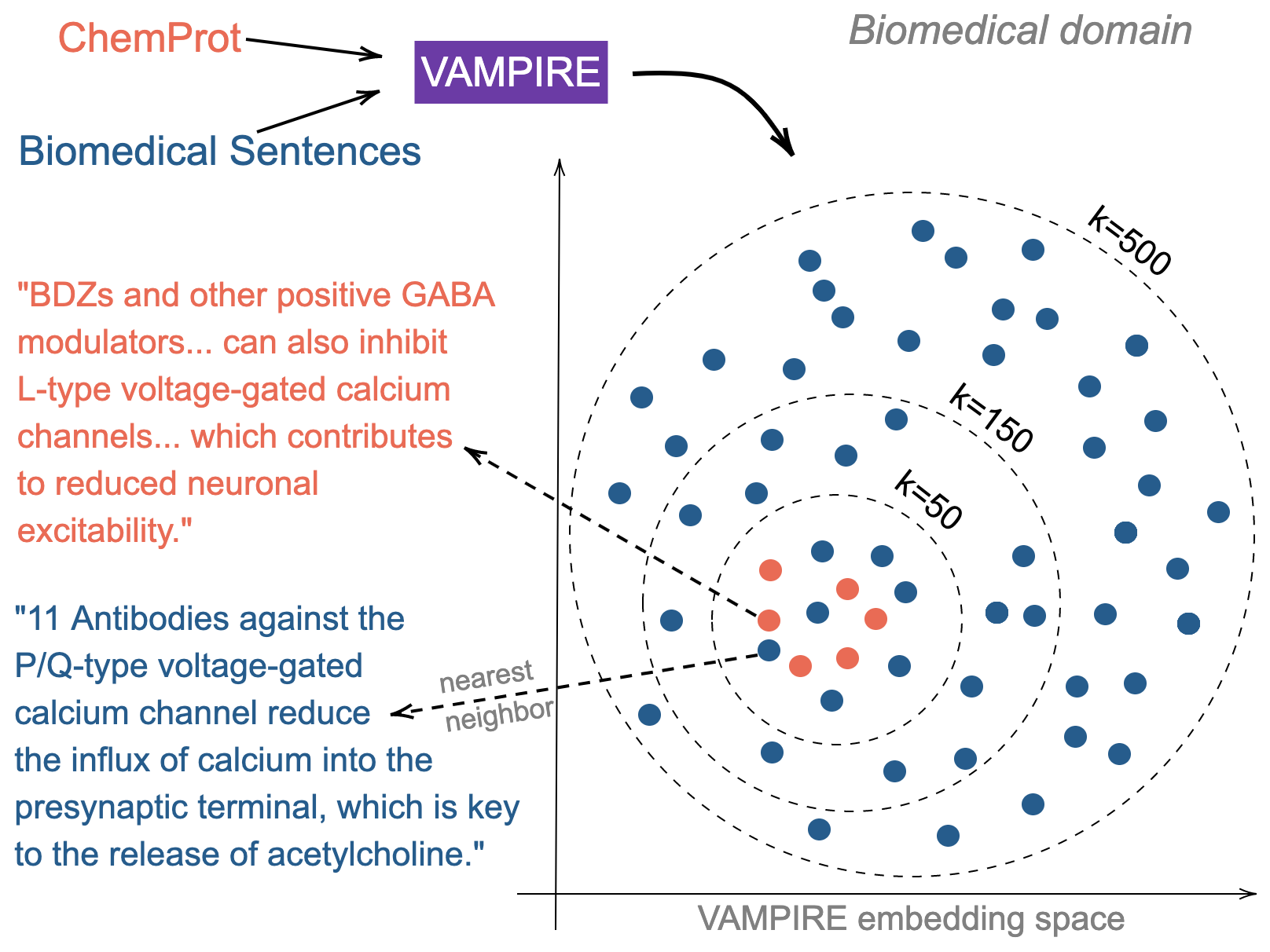}
     \caption{An illustration of automated data selection (\S\ref{sec:selection}). We map unlabeled \chemprot and 1M \med sentences to a shared vector space using the \vampire model trained on these sentences. Then, for each \chemprot sentence, we identify $k$ nearest neighbors, from the \med domain.}
     \label{fig:selection_intuition}
\end{figure}

 \begin{table}[t]
\centering

\resizebox{\columnwidth}{!}{
\begin{tabular}{cccc}
\toprule
\multirow{2}{*}{Pretraining}  & \multicolumn{2}{c}{\med} & \cs  \\
&   \chemprot  & \rct-500         & \arccite  \\
\midrule
\roberta            &   81.9$_{1.0}$ &	79.3$_{0.6}$ &	63.0$_{5.8}$   \\
\tapt &    82.6$_{0.4}$ & 79.8$_{1.4}$ &	67.4$_{1.8}$    \\
\midrule[0.03em]
\random &  81.9$_{0.6}$  &	80.6$_{0.4}$  &	 69.7$_{3.4}$      \\
\knn{50} &     83.3$_{0.7}$ &	80.8$_{0.6}$ &	70.7$_{2.8}$ \\
\knn{150} &     83.2$_{0.6}$     & 81.2$_{0.8}$ & 73.3$_{2.7}$ \\
\knn{500} &   83.3$_{0.7}$  & 81.7$_{0.4}$ & \bf 75.5$_{1.9}$	 \\
\midrule 
    \dapt & \bf 84.2$_{0.2}$ & \bf 82.5$_{0.5}$ & 75.4$_{2.5}$ \\
\bottomrule
\end{tabular}
}
\caption{Mean test set micro-$F_1$ (for \chemprot and \rct) and macro-$F_1$ (for \arccite), across five random seeds, with standard deviations as subscripts, comparing \random (with 50 candidates) and \knn{k} selection.
Neighbors of the task data are selected from the domain data.
% \ana{Mark: replace with bar plot}
% Tried this notation to normalize random and nearest neighbors. \nascomment{see comment in text about abbreviations}
 }
\label{tab:test_selection_results}
\end{table}

\subsection{Automated Data Selection for \tapt}
\label{sec:selection}

Consider a low-resource scenario without access to large amounts of unlabeled data to adequately benefit from \tapt, as well as absence of computational resources necessary for \dapt (see Table \ref{tab:speed} for details of computational requirements for different pretraining phases). 
We propose simple unsupervised methods to retrieve unlabeled text that aligns with the task distribution, from a large in-domain corpus. 
Our approach finds task-relevant data from the domain by embedding text from both the task and domain in a shared space, then selects candidates from the domain based on queries using the task data.  
Importantly, the embedding method must be lightweight enough to embed possibly millions of sentences in a reasonable time. 

% \begin{figure}[t]
%      \centering
%      \includegraphics[width=3.0in]{figs/selection_fig_ana.png}
%      \caption{An illustration of automated data selection for TAPT with VAMPIRE. 
%      } 
%      \label{fig:selection_intuition}
%  \end{figure}

Given these constraints, we employ \vampire (\citealp{gururangan2019variational}; Figure \ref{fig:selection_intuition}), a lightweight bag-of-words language model. 
We pretrain \vampire on a large deduplicated\footnote{We deduplicated this set to limit computation, since different sentences can share neighbors.} sample of the domain (1M sentences) to obtain embeddings of the text from both the task and domain sample.  
We then select $k$ candidates of each task sentence from the domain sample, in embeddings space.
Candidates are selected (i) via nearest neighbors selection (\knn{k})\footnote{We use a flat search index with cosine similarity between embeddings with the \texttt{FAISS} \cite{JDH17} library.}, or (ii) randomly (\random).
% The central idea here is that instead of relying on all the data from a large unlabeled corpus for the domain, we want to exploit cheaper alternatives.
We continue pretraining \roberta on this augmented corpus with both the task data (as in \tapt) as well as the selected candidate pool. 
% More concretely, we first sample a large set (1M in our experiments) of unlabeled sentences % \footnote{We sample sentences from the corpus because the tasks we evaluate on are sentence-level. With a low-resource document-level task, sampling in-domain documents might be more useful.} 
% from an in-domain corpus, such as \cs or \med. 
% In the experiments we outline in this section, we always sample 1M sentences from the domain.

% Next, we require a method for embedding texts from the task and the domain in a shared space. 

% \footnote{Future work could experiment with larger amounts of unlabeled data, which could warrant quantization or other compression techniques to perform the nearest neighbors search queries.} 

% \nascomment{I don't like nbr as an abbreviation for neighbor.  I would use (e.g.) 150\textsc{nn} for nearest and something else for random (can't give advice because i'm still confused about what random nn means}

\paragraph{Results} Results in Table \ref{tab:test_selection_results} show that \knn{k} outperforms \tapt for all cases. 
\random is generally worse than \knn{k}, but within a standard deviation arising from 5 seeds for \rct and \arccite.
As we increase $k$, \knn{k} performance steadily increases, and approaches that of \dapt.
Appendix \ref{sec:appendix_knn} shows examples of nearest neighbors of task data.
% More comprehensive tuning of $k$, a better document embedder, or larger sample of the domain may improve the results of \knn{k} against random selection.
% Moreover, \arccite performance decreases beyond sentene  a certain $k$ hinting that after a point, we probably start collecting task-irrelevant samples, which might not be informative for adaptation.
% Table \ref{tab:speed} shows that the x
% We also compare with a baseline which randomly samples examples from the domain, to control for the effect of just ``more data'' vs. selection.
% This baseline while worse than our \knn, comes quite close.
Future work might consider a closer study of \knn{k}, more sophisticated data selection methods, and the tradeoff between the diversity and task relevance of selected examples.

% Increasing $k$ likely introduces more diverse samples into the dataset, which improves the generalization performance of \roberta. 
%\nascomment{this is not very clear; what I see in the table right now is that augmenting \tapt with more data, drawn in this way, is quite helpful (hopefully more so than `random,' which is a control for the effect of just ``more data'' vs. selection -- that needs to be explained)} 
%\ana{This. It seems like we won't have better results than the random baseline, so let's be honest here and finish with the next sentence.} 
%\nascomment{missing:  how much more expensive are these data augmentation strategies, above and beyond the cost of tapt?  I feel like part of the motivation here is that this is cheap, but we haven't given any evidence or stated that claim explicitly!}
% Here we have explored with one simple method for augmenting \tapt; future work may reveal approaches that are even more efficient, more effective, or both.
%\swabha{Rewrote this section and tried addressing all comments.}

%\paragraph{Implications}

%\suchin{Here we can talk about the fact that we've only explored one method for selection, many others exist, but data selection generally has interesting future work for domain adaption under low resource situations} \nascomment{keep it short and just add as a sentence to the above, doesn't need a bold call-out:} 

\subsection{Computational Requirements}
\label{sec:speed}

The computational requirements for all our adaptation techniques on \rct-500 in the \med domain in Table \ref{tab:speed}.
\tapt is nearly 60 times faster to train than \dapt on a single v3-8 TPU and storage requirements for \dapt on this task are 5.8M times that of \tapt.
Our best setting of \dapt+ \tapt amounts to three phases of pretraining, and at first glance appears to be very expensive.
However, once the LM has been adapted to a broad domain, it can be reused for multiple tasks within that domain, with only a single additional \tapt phase per task.
% \nascomment{I cut here, I think it was a bit too bold}
While \oracle tends to achieve the best cost-benefit ratio in this comparison, one must also take into account the cost of curating large in-domain data. 
Automatic methods such as \knn{k} are much cheaper than \dapt.

\begin{table}[t]
    \small
    \centering
    \resizebox{\columnwidth}{!}{

    \hspace*{-2.5mm}
    \begin{tabular}{crrrr}
        \toprule
        Pretraining & Steps  & Docs. & Storage & $F_1$ \\
        \midrule
        \roberta       & -      & - & - & 79.3$_{0.6}$\\
        \midrule[0.03em]
        \tapt           &  0.2K  & 500 & 80KB  &79.8$_{1.4}$ \\
        \knn{50}  &  1.1K  & 24K & 3MB  &80.8$_{0.6}$\\
        \knn{150} &  3.2K  & 66K & 8MB  &81.2$_{0.8}$\\
        \knn{500} &  9.0K  & 185K & 24MB &81.7$_{0.4}$\\
        \oracle        &  8.8K  & 180K & 27MB  & \textbf{83.4$_{0.3}$}\\
        \dapt          & 12.5K  & 25M &  47GB & 82.5$_{0.5}$\\
        \dapt + \tapt  & 12.6K  & 25M &  47GB & 83.0$_{0.3}$\\
         \bottomrule
    \end{tabular}
    }
\caption{Computational requirements for adapting to the \rct-500 task, comparing \dapt (\sect{sec:dapt}) and the various \tapt modifications described in \sect{sec:tapt} and \sect{sec:more-tapt}.
}
    \label{tab:speed}
\end{table}

\section{Related Work}
\label{sec:related}

\paragraph{Transfer learning for domain adaptation}

Prior work has shown the benefit of continued pretraining in domain 
\citep{alsentzer-etal-2019-publicly, chakrabarty-etal-2019-imho, Lee2019BioBERTAP}.\footnote{In contrast, \citet{peters-etal-2019-tune} find that the Jensen-Shannon divergence on term distributions between \bert's pretraining corpora and each \textsc{MultiNLI} domain \cite{williams-etal-2018-broad} does not predict its performance, though this might be an isolated finding specific to the MultiNLI dataset.
% \nascomment{note to self - check here} 
% \swabha{Not sure this is worth including in the main text, moving to footnote. It seems to be a solitary result, compared to the vast amount of results supporting the importance of domains. Also it seems to be more about MultiNLI, which we know it's an unreliable dataset, since changing the domain does not seem to have any effect on the task. Acc. to the paper, ``At least for this task, the distance of the source and target domains does not seem to have a major impact on the adaptation performance.'' Moreover, we don't use MNLI in our experiments either!}
}
We have contributed further investigation of the effects of a shift between a large, diverse pretraining corpus and target domain on task performance.
Other studies \citep[e.g.,][]{Huang2019ClinicalBERTMC}
have trained language models (LMs) in their domain of interest, from scratch.
In contrast, our work explores multiple domains, and is arguably more cost effective, since we continue pretraining an already powerful LM. 

\paragraph{Task-adaptive pretraining} 
Continued pretraining of a LM on the unlabeled data of a given task (\tapt) has been show to be beneficial for end-task performance \cite[e.g.\ in][]{howard-ruder-2018-universal, Phang2018SentenceEO, Sun2019HowTF}. 
In the presence of \textit{domain shift} between train and test data distributions of the same task, domain-adaptive pretraining (\dapt) is sometimes used to describe what we term \tapt \cite{logeswaran-etal-2019-zero, han_eisenstein_emnlp_2019}. 
Related approaches include language modeling as an auxiliary objective to task classifier fine-tuning \citep{chronopoulou-etal-2019-embarrassingly, radford2018improving} or consider simple syntactic structure of the input while adapting to task-specific data \cite{swayamdipta2019shallow}. 
%\citet{Sun2019HowTF} explore domain- and task-adaptive pretraining across three unconventional domains (i.e., sentiment, question, topic).
We compare \dapt and \tapt as well as their interplay with respect to dataset size for continued pretraining (hence, expense of more rounds of pretraining), relevance to a data sample of a given task, and transferability to other tasks and datasets.
See Table \ref{tab:related_work} in Appendix \sect{sec:related_work_overview} for a summary of multi-phase pretraining strategies from related work.

\begin{table}[]
\resizebox{\columnwidth}{!}{

%\begin{tabular}{cm{1.5cm}m{1.5cm}m{1.5cm}}
\begin{tabular}{cccc}
\toprule
& \multicolumn{3}{c}{\bf Training Data} \\
\cmidrule{2-4} 
%  \diagbox{Model}{Trained On}
 & \makecell{Domain\\(Unlabeled)}
 & \makecell{Task\\(Unlabeled)}
 & \makecell{Task\\(Labeled)} \\
\midrule
\roberta &  &  & \checkmark \\
\dapt & \checkmark & & \checkmark\\
\tapt &  & \checkmark & \checkmark \\
\dapt + \tapt & \checkmark & \checkmark & \checkmark\\
\knn{k} & (Subset) & \checkmark & \checkmark \\
\oracle &  & (Extra) & \checkmark \\
\bottomrule
\end{tabular}
}
\caption{Summary of strategies for multi-phase pretraining explored in this paper. }
\label{tab:apt_desc}
\end{table}

\paragraph{Data selection for transfer learning} 
Selecting data for transfer learning has been explored in NLP \cite[among others]{moore-lewis-2010-intelligent, ruder-plank-2017-learning,zhang-etal-2019-curriculum}. % more: bhatt-etal-2016-cross,rehbein_bildhauer_2017,
\citet{dai-etal-2019-using} focus on identifying the most suitable corpus to pretrain a LM from scratch, for a single task: NER, whereas we select relevant \emph{examples} for various tasks in \S\ref{sec:selection}.
% (in contrast to , in our case)
% In contrast, % we propose an approach to
% our approach efficiently selects  from target domain data that are closer to task data distribution, for continued pretraining. 
Concurrent to our work, \citet{Aharoni2020UnsupervisedDC} propose data selection methods for NMT based on cosine similarity in embedding space, using \textsc{DistilBERT} \cite{Sanh2019DistilBERTAD} for efficiency.
In contrast, we use \vampire, and focus on augmenting \tapt data for text classification tasks.
\citet{Khandelwal2019GeneralizationTM} introduced $k$\textsc{NN}-LMs that allows easy domain adaptation of pretrained LMs by simply adding a datastore per domain and no further training; an alternative to integrate domain information in an LM.  
% We draw attention to the importance of domain- and task-specific data, and their work provides an effective approach 
Our study of human-curated data \sect{sec:oracle} is related to \textit{focused crawling} \cite{Chakrabarti1999FocusedCA} for collection of suitable data, especially with LM reliance \cite{remus-biemann-2016-domain}.

\paragraph{What is a domain?} 
Despite the popularity of domain adaptation techniques, most research and practice seems to use an intuitive understanding of domains.
A small body of work has attempted to address this question \citep[among others]{lee2001genres, Eisenstein2014DiffusionOL, van-der-wees-etal-2015-whats, plank_2016, ruder-etal-2016-towards}. 
For instance, \citet{Aharoni2020UnsupervisedDC} define domains by implicit clusters of sentence representations in pretrained LMs. 
% We reflect on this question in the context of , and show that these two approaches complement each other. 
Our results show that \dapt and \tapt complement each other, which suggests a 
% non-discrete 
spectra of domains defined around tasks
% and provenance, 
at various levels of granularity (e.g., Amazon reviews for a specific product, 
all Amazon reviews, all reviews on the web, the web).
% \nascomment{reworked here}

\section{Conclusion}
\label{sec:conclusion}

We investigate several variations for adapting pretrained LMs to domains and tasks within those domains, summarized in Table \ref{tab:apt_desc}. 
Our experiments reveal that even a model of hundreds of millions of parameters struggles to encode the complexity of a single textual domain, let alone all of language. 
We show that pretraining the model towards a %\swabha{specific domain and a}
specific task or small corpus can provide significant benefits.
Our findings suggest it may be valuable to complement work on ever-larger LMs with parallel efforts to identify and use domain- and task-relevant corpora to specialize models.
While our results demonstrate how these approaches can improve \roberta, a powerful LM, the approaches we studied are general enough to be applied to any pretrained LM. 
Our work points to numerous future directions, such as better data selection for \tapt, efficient adaptation large pretrained language models to distant domains, and building reusable language models after adaptation. 

% We investigate the adaptation of large-scale pretrained LMs into several domains and tasks within each.
% All our methods build on and improve a powerful LM, \roberta; however, are general enough to be applied to any pretrained LM.

% How to better select data for TAPT under low-resource environments?
% How to efficiently adapt models to different corpora?
% How to alleviate the negative cross-task transfer associated with TAPT?
%  \swabha{Want to add that phrase as a general recommendation to readers.}

%Our code and data\footnote{\ana{Be careful with the data, we cannot release RealNews for example.} Ana: can you specify what the restrictions with \realnews are?}, will be made publicly available upon publication.
% Our pretrained models as well as code, will be made publicly available upon publication. 
% \nascomment{make this point in the intro, too}

\section*{Acknowledgments}
% Arman for data for MLM-roberta. 
 The authors thank
 Dallas Card, Mark Neumann, Nelson Liu, Eric Wallace, members of the AllenNLP team, and anonymous reviewers for helpful feedback, and Arman Cohan for providing data. This research was supported in part by the Office of Naval Research under the MURI grant N00014-18-1-2670. TPU machines for conducting experiments were provided
by Google.

\bibliography{references}
\bibliographystyle{acl_natbib}

\clearpage

\appendix

\begin{table*}[!h]
\resizebox{\textwidth}{!}{
\begin{tabular}{l>{\raggedright\arraybackslash}m{3.8cm}>{\raggedright\arraybackslash}m{3cm}>{\raggedright\arraybackslash}m{2.8cm}>{\centering\arraybackslash}m{1cm}>{\centering\arraybackslash}m{1cm}>{\centering\arraybackslash}m{1.1cm}>{\centering\arraybackslash}m{1cm}>{\centering\arraybackslash}m{1cm}}
\toprule
& \makecell[l]{\textsc{DAPT} Domains \\(if applicable)} & Tasks & Model & \dapt & \tapt & \dapt + \tapt & \knn{k} & \oracle\\
\midrule
\textbf{This Paper} & biomedical \& computer science papers, news, reviews & 8 classification tasks & \roberta & \checkmark & \checkmark & \checkmark  & \checkmark  & \checkmark  \\
\midrule
\citet{Aharoni2020UnsupervisedDC} & \makecell[c]{-} & NMT & \textsc{DistilBERT} + Transformer NMT & - &  - & - & similar & - \\ 
\midrule
\citet{alsentzer-etal-2019-publicly} & clinical text & NER, NLI, de-identification & \textsc{(Bio)BERT} & \checkmark &  - & - & - & - \\ 
\midrule
\citet{chakrabarty-etal-2019-imho} & opinionated claims from Reddit & claim detection & \textsc{ULMFiT} & \checkmark & \checkmark & - & - & - \\ 
\midrule 
\citet{chronopoulou-etal-2019-embarrassingly} & \makecell[c]{-} & 5 classification tasks & \textsc{ULMFiT}$^{\dagger}$ & - & similar & - & - & -\\
\midrule
\citet{han_eisenstein_emnlp_2019} & \makecell[c]{-} &  NER in historical texts  & \textsc{ELMo}, \textsc{BERT} & - & \checkmark & - & - & - \\ 
\midrule
\citet{howard-ruder-2018-universal} & \makecell[c]{-} &  6 classification tasks & \textsc{ULMFiT} & -  & \checkmark & - & - & - \\
\midrule
\citet{Khandelwal2019GeneralizationTM} & \makecell[c]{-} & language modeling  & Transformer LM & -  & - & - & similar & - \\
\midrule
\citet{Lee2019BioBERTAP} & biomedical papers & NER, QA, relation extraction & \textsc{BERT} & \checkmark & - & - & - & -\\ 
\midrule
\citet{logeswaran-etal-2019-zero} & \makecell[c]{-} &  zero-shot entity linking in Wikia & \textsc{BERT} & -  & \checkmark & - & - & - \\
\midrule
\citet{Mitra2019ExploringWT} & \makecell[c]{-} & commonsense QA & \textsc{BERT} & - & \checkmark & - & - & - \\
\midrule
\citet{Phang2018SentenceEO} & \makecell[c]{-} & \textsc{GLUE} tasks & \textsc{ELMo, BERT, GPT} & - & \checkmark & - & - & - \\
\midrule
\citet{radford2018improving} & \makecell[c]{-} & NLI, QA, similarity, classification & \textsc{GPT} & - & similar & - & - & -\\
\midrule
\citet{Sun2019HowTF} &  sentiment, question, topic & 7 classification tasks & \textsc{BERT} & \checkmark & \checkmark & - & - & - \\
\midrule
\citet{swayamdipta2019shallow} & \makecell[c]{-} & NER, parsing, classification & \textsc{ELMo} & - & similar & - & - & -\\
\midrule 
\citet{xu-etal-2019-bert} & reviews & RC, aspect extract., sentiment classification & \textsc{BERT} & \checkmark & \checkmark & \checkmark & - & - \\
\midrule
\citet{Xu2019ReviewCR} & restaurant reviews, laptop reviews & conversational RC & \textsc{BERT} & \checkmark & \checkmark & - & - & - \\
\bottomrule
\end{tabular}
}
\caption{Overview of prior work across strategies for continued pre-training summarized in Table \ref{tab:apt_desc}. \textsc{ULMFiT} is pretrained on English Wikipedia; \textsc{ULMFiT}$^{\dagger}$ on English tweets; \textsc{ELMo} on the \textsc{1BWordBenchmark} \cite[newswire; ][]{1BWordBenchmark}; \textsc{GPT} on \textsc{BookCorpus}; \textsc{BERT} on English Wikipedia and \textsc{BookCorpus}. In comparison to these pretraining corpora, \roberta's pretraining corpus is substantially more diverse (see Appendix \sect{sec:roberta_pretraining}). }
\label{tab:related_work}
\end{table*}

\section*{Appendix Overview}

In this supplementary material, we provide: (i) additional information for producing the results in the paper, and (ii) results that we could not fit into the main body of the paper.  

\smallskip

\noindent\textbf{Appendix \ref{sec:related_work_overview}.} A tabular overview of related work described in Section \sect{sec:related}, a description of the corpus used to train \roberta in \citet{Liu2019RoBERTaAR}, and references to the state of the art on our tasks.

\smallskip

\noindent\textbf{Appendix \ref{sec:training}.} Details about the data preprocessing, training, and implementation of domain- and task-adaptive pretraining.

\smallskip

\noindent\textbf{Appendix \ref{sec:dev_set_results}.} Development set results.

\smallskip

\noindent\textbf{Appendix \ref{sec:appendix_domain_transfer}.} Examples of domain overlap. 

\smallskip

\noindent\textbf{Appendix \ref{sec:mlm_study_appendix}.} The cross-domain masked LM loss and reproducibility challenges. 

\smallskip

\noindent\textbf{Appendix \ref{sec:appendix_knn}.} Illustration of our data selection method and examples of nearest neighbours.

\section{Related Work}
\label{sec:related_work_overview}
Table \ref{tab:related_work} shows which of the strategies for continued pretraining have already been explored in the prior work from the Related Work (\sect{sec:related}). As evident from the table, our work compares various strategies as well as their interplay using a  pretrained language model trained on a much more heterogeneous pretraining corpus.

\subsection{\roberta's Pretraining Corpus}
\label{sec:roberta_pretraining}

\roberta was trained on data from \books \cite{Zhu2015AligningBA},\footnote{\url{https://github.com/soskek/bookcorpus}} \wiki,\footnote{\url{https://github.com/google-research/bert}} a portion of the \ccnews dataset \cite{nagel2016ccnews},\footnote{\url{https://github.com/fhamborg/news-please}} \textsc{OpenWebText} corpus of Web content extracted from URLs shared on Reddit \cite{Gokaslan2019OpenWeb},\footnote{\url{https://github.com/jcpeterson/openwebtext}} and a subset of CommonCrawl that it is said to resemble the ``story-like'' style of \textsc{Winograd} schemas \citep[\textsc{STORIES}; ][]{Trinh2018ASM}.\footnote{\url{https://github.com/tensorflow/models/tree/master/research/lm_commonsense}}

\subsection{State of the Art}
\label{sec:sota}

In this section, we specify the models achieving state of the art on our tasks. See the caption of Table \ref{tab:main} for the reported performance of these models. For \arccite, that is \textsc{SciBERT} \cite{Beltagy2019SciBERTAP}, a \bert-base model for trained from scratch on scientific text.  For \chemprot and \sciie, that is \textsc{S2ORC-BERT} \cite{lo2019gorc}, a similar model to \textsc{SciBERT}. For \ag and \imdb, XLNet-large, a much larger model.  For \rct, \citet{cohan-etal-2019-pretrained}.  For \hp, \textsc{Longformer}, a modified Transformer language model for long documents \cite{Beltagy2020Longformer}. \citet{thongtan-phienthrakul-2019-sentiment} report a higher number ($97.42$) on \imdb, but they train their word vectors on the test set. Our baseline establishes the first benchmark for the \helpful dataset. 

\section{Experimental Setup}
\label{sec:training}

\paragraph{Preprocessing for \dapt}

The unlabeled corpus in each domain was pre-processed prior to language model training.
Abstracts and body paragraphs from biomedical and computer science articles were used after sentence splitting using \texttt{scispaCy} \citep{Neumann_2019}.  We used summaries and full text of each news article, and the entire body of review from Amazon reviews. For both news and reviews, we perform sentence splitting using \texttt{spaCy} \citep{spacy2}. 

% \todo{Can we emphasize that we do LM training on a document-level, not sentence-level?} \kyle{What does this mean?  Is it that we fit contiguous sentences into blocks of 512-length sequences? }
% \paragraph{Downstream task datasets}
% To simulate a setting in which we have large amounts of curated data for macro task-adaptive fine-tuning in the medical domain, we downsample the training data of the \rct dataset to 500 examples, and treat the rest of the training data as unlabeled.
% \todo{How did we decide which datasets to down-sample for a low-resource study? }
% \todo{Describe each classification task in a sentence? Just input and output.}

\paragraph{Training details for \dapt} We train \roberta on each domain for 12.5K steps. We focused on matching all the domain dataset sizes (see Table \ref{tab:domain_datasets}) such that each domain is exposed to the same amount of data as for 12.5K steps it is trained for. \amazon contain more documents, but each is shorter. We used an effective batch size of 2048 through gradient accumulation, as recommended in \citet{Liu2019RoBERTaAR}. See Table \ref{tab:roberta_adaptive_pretraining_hyperparameters} for more hyperparameter details.

\paragraph{Training details for \tapt} We use the same pretraining hyperparameters as \dapt, but we artificially augmented each dataset for \tapt by randomly masking different tokens across epochs, using the masking probability of 0.15. Each dataset was trained for 100 epochs.  For tasks with less than 5K examples, we used a batch size of 256 through gradient accumulation.  See Table \ref{tab:roberta_adaptive_pretraining_hyperparameters} for more hyperparameter details.

\paragraph{Optimization} We used the Adam optimizer \cite{kingma2014adam}, a linear learning rate scheduler with $6$\% warm-up, a maximum learning rate of $0.0005$. When we used a batch size of 256, we used a maximum learning rate of $0.0001$, as recommended in \citet{Liu2019RoBERTaAR}. We observe a high variance in performance between random seeds when fine-tuning \roberta to \hp, because the dataset is extremely small. To produce final results on this task, we discard and resample degenerate seeds. We display the full hyperparameter settings in  Table \ref{tab:roberta_adaptive_pretraining_hyperparameters}.

\paragraph{Implementation} Our LM implementation uses the \texttt{HuggingFace} \texttt{transformers} library \citep{wolf2019huggingfaces}\footnote{\url{https://github.com/huggingface/transformers}} and \texttt{PyTorch} \texttt{XLA} for TPU compatibility.\footnote{\url{https://github.com/pytorch/xla}}  
Each adaptive pretraining experiment was performed on a single v3-8 TPU from Google Cloud.\footnote{\url{http://github.com/allenai/tpu-pretrain}} For the text classification tasks, we used \texttt{AllenNLP} \cite{gardner2018allennlp}. Following standard practice \cite{devlin-etal-2019-bert} we pass the final layer \texttt{[CLS]} token representation to a task-specific feedforward layer for prediction.
% We use a batch size of 2048 and the maximum sequence length of 512 word pieces. 
% We use the AdamW optimizer, with a learning rate of \todo{?}. 
% We train for \todo{?} epochs, checkpointing the model after each epoch. 

\section{Development Set Results}
\label{sec:dev_set_results}

Adhering to the standards suggested by \citet{dodge2019show} for replication, we report our development set results in Tables \ref{tab:dev_main},
\ref{tab:transfer_dev_results}, and \ref{tab:dev_micro_vs_macro_task_adaptive_results}.

\section{Analysis of Domain Overlap}
\label{sec:appendix_domain_transfer}

In Table \ref{tab:more-domain-overlap} we display additional examples that highlight the overlap between \imdb reviews and \realnews articles, relevant for analysis in \sect{sec:domain_boundaries}. 

\section{Analysis of Cross-Domain Masked LM Loss}
\label{sec:mlm_study_appendix}
\begin{table*}[t]
\center
\small
\begin{tabular}{llccccc}
\toprule
 && \multicolumn{5}{c}{Data Sample Unseen During \dapt}\\
 \cmidrule{3-7} 
 & & PT & \med & \cs & \news & \reviews \\
 \midrule
& \roberta & \textbf{1.19} & 1.32 & 1.63 & \textbf{1.08} & 2.10 \\
\ldelim\{{4}{10mm}[\dapt]  & \med & 1.63 & \textbf{0.99} & 1.63 & 1.69 & 2.59 \\
& \cs & 1.82 & 1.43 & \textbf{1.34} & 1.92 & 2.78 \\
& \news & 1.33 & 1.50 & 1.82 & 1.16 & 2.16 \\
& \reviews & 2.07 & 2.23 & 2.44 & 2.27 & \textbf{1.93}    \\
\bottomrule
\end{tabular}
\caption{\roberta's (row 1) and domain-adapted \roberta's (rows 2--5) masked LM loss on randomly sampled held-out documents from each domain (lower implies a better fit). \PT denotes a sample from sources similar to \roberta's pretraining corpus. The lowest masked LM for each domain sample is boldfaced.}
\label{tab:masked_lm_study}
\end{table*}

%\begin{table*}[t]
%\center
%\begin{tabular}{llccccc}
%\toprule
% && \multicolumn{5}{c}{Data Sample Unseen During \dapt}\\
% \cmidrule{3-7} 
% & & PT & \med & \cs & \news & \reviews \\
% \midrule
%& \roberta & 1.66            & 1.85       & 2.30      & \textbf{1.51}        & \textbf{2.39} \\
%\ldelim\{{4}{10mm}[\dapt]  & \med & 2.55  & \textbf{1.54}       & 2.45      & 2.52        & 3.85 \\
%& \cs & 2.83            & 2.17       & \textbf{2.05}      & 2.86        & 4.13           \\
%& \news & 2.12            & 2.24       & 2.70      & 1.78        & 3.26           \\
%& \reviews & 3.11            & 3.28       & 3.59      & 3.36        & 3.34    \\
%\bottomrule
%\end{tabular}
%\caption{\roberta's and domain-adapted \roberta's masked LM loss on randomly sampled held-out documents from each domain (lower implies a better fit). PT denotes a sample of \roberta's pretraining corpus.}
%\label{tab:masked_lm_study}
%\end{table*}
% \swabha{We have to be very careful what we write about this, even if it is in the appendix, so that readers get the most out of the appendix.
% Some of the things written here are directly contradicting our results.}

In Section \sect{sec:dapt_experiments}, we provide \roberta's masked LM loss before and after \dapt. We display cross-domain masked-LM loss in Table \ref{tab:masked_lm_study}, where we evaluate masked LM loss on text samples in other domains after performing \dapt. 

We observe that the cross-domain masked-LM loss mostly follows our intuition and insights from the paper, i.e.\ \roberta's pretraining corpus and \news are closer, and \med to \cs (relative to other domains). 
However, our analysis in \sect{sec:domain_boundaries} illustrates that \reviews and \news also have some similarities. 
This is supported with the loss of \roberta that is adapted to \news, calculated on a sample of \reviews. 
However, \roberta that is adapted to \reviews results in the highest loss for a \news sample. 
This is the case for all domains. 
One of the properties that distinguishes \reviews from all other domains is that its documents are significantly shorter. 
In general, we find that cross-\dapt masked-LM loss can in some cases be a noisy predictor of domain similarity.
% At the moment we can not firmly confirm that this property is what causes a relatively higher masked LM loss of \roberta adapted to \reviews on samples of other domains. 

\section{$k$-Nearest Neighbors Data Selection}
\label{sec:appendix_knn}
In Table \ref{tab:knn_ex}, we display nearest neighbor documents in the \med domain identified by our selection method, on the \rct dataset.

\begin{table*}[t!]
    \centering
    \small

    \begin{tabular}{cc}
      \toprule
      \textbf{Computing Infrastructure} & Google Cloud v3-8 TPU\\ 
      \midrule
      \textbf{Model implementations} & \url{https://github.com/allenai/tpu_pretrain}\\
      \bottomrule
    \end{tabular}
    
    \vspace{3mm}\begin{tabular}{cc}
        \toprule
        \textbf{Hyperparameter} & \textbf{Assignment}  \\
        \midrule
        number of steps & 100 epochs (\tapt) or 12.5K steps (\dapt) \\
        \midrule
        batch size & 256 or 2058 \\
        \midrule
        maximum learning rate & 0.0001 or 0.0005 \\
        \midrule
        learning rate optimizer & Adam \\
        \midrule
        Adam epsilon & 1e-6 \\
        \midrule
        Adam beta weights & 0.9, 0.98\\
        \midrule
        learning rate scheduler & None or warmup linear \\
        \midrule
        Weight decay & 0.01 \\
        \midrule
        Warmup proportion & 0.06 \\
        \midrule
        learning rate decay & linear \\
        \bottomrule
    \end{tabular}
    
    % \vspace{3mm}\begin{tabular}{cc}
    %   \toprule
    %   \textbf{Dataset} & \textbf{\VAMPIRE~NPMI} \\ 
    %   \midrule
    %   \imdb & 0.131\\ 
    %   \midrule
    %   \ag & 0.224  \\ 
    %   \midrule
    %   \yahoo & 0.475 \\ 
    %   \midrule
    %   \hatespeech & 0.139 \\ 
    %   \bottomrule
    % \end{tabular}
    
    \caption{Hyperparameters for domain- and task- adaptive pretraining.} 
    \label{tab:roberta_adaptive_pretraining_hyperparameters}
\end{table*}

\begin{table*}[t!]
    \centering
    \small

    \begin{tabular}{cc}
      \toprule
      \textbf{Computing Infrastructure} & Quadro RTX 8000 GPU \\
      \midrule
      \textbf{Model implementation} & \url{https://github.com/allenai/dont-stop-pretraining}\\
      \bottomrule
    \end{tabular}

    \vspace{3mm}\begin{tabular}{cc}
        
        \toprule
        \textbf{Hyperparameter} & \textbf{Assignment}  \\
        \midrule
        number of epochs & 3 or 10 \\
        \midrule
        patience & 3  \\
        \midrule
        batch size & 16 \\
        \midrule
        learning rate & 2e-5\\
        \midrule
        dropout & 0.1 \\
        \midrule
        feedforward layer & 1 \\
        \midrule
        feedforward nonlinearity & tanh \\
        \midrule
        classification layer & 1 \\
        \bottomrule
    \end{tabular}
    
    % \vspace{3mm}\begin{tabular}{cc}
    %   \toprule
    %   \textbf{Dataset} & \textbf{\VAMPIRE~NPMI} \\ 
    %   \midrule
    %   \imdb & 0.131\\ 
    %   \midrule
    %   \ag & 0.224  \\ 
    %   \midrule
    %   \yahoo & 0.475 \\ 
    %   \midrule
    %   \hatespeech & 0.139 \\ 
    %   \bottomrule
    % \end{tabular}
    
    \caption{Hyperparameters for \roberta text classifier. } 
    \label{tab:roberta_text_classification_hyperparameters}
\end{table*}

\begin{table*}[t]
\centering
\small
% \resizebox{0.9\textwidth}{!}{
\begin{tabular}{llcccc}
\toprule
 &  & & \multicolumn{3}{c}{Additional Pretraining Phases}\\
\cline{4-6}
\textbf{Domain} &   \textbf{Task} 
& \bf \roberta & \textbf{\dapt} & \textbf{\tapt} & \textbf{\dapt + \tapt} \\ %& \boldmath{\lnot}\textbf{\dapt}\\
\midrule 
\multirow{2}{*}{\med} &   \chemprot 
& 83.2$_{1.4}$& \bf 84.1$_{0.5}$& 83.0$_{0.6}$& \bf 84.1$_{0.5}$\\ %& 80.6$_{1.2}$\\
& $^\dagger$\rct  
& 88.1$_{0.05}$ & \bf 88.5$_{0.1}$ &  88.3$_{0.1}$ & \bf 88.5$_{0.1}$ \\ %& 72.7$_{0.7}$\\ 
\midrule[0.03em]
\multirow{2}{*}{\cs}& \arccite 
& 71.3$_{2.8}$& 73.2$_{1.5}$ & 73.2$_{3.6}$&  \bf 78.6$_{2.9}$\\ %& 67.7$_{5.7}$\\
& \sciie
& 83.8$_{1.1}$& \bf 88.4$_{1.7}$& 85.9$_{0.8}$ &  88.0$_{1.3}$\\ %& 80.0$_{3.0}$\\
\midrule[0.03em]
\multirow{2}{*}{\news}& \hp 
% 82.1
& 84.0$_{1.5}$& 79.1$_{3.5}$& \bf 82.7$_{3.3}$  & 80.8$_{2.3}$\\ %& 79.0$_{4.1}$\\
& $^\dagger$\ag 
& 94.3$_{0.1}$& 94.3$_{0.1}$&  94.7$_{0.1}$& \bf 94.9$_{0.1}$\\ %& 93.4$_{0.2}$\\
\midrule[0.03em]
\multirow{2}{*}{\reviews} & $^\dagger$\helpful
& 65.5$_{3.4}$& 66.5$_{1.4}$& 69.2$_{2.4}$&  \bf 69.4$_{2.1}$\\ %&  66.9$_{3.2}$\\
& $^\dagger$\imdb 
& 94.8$_{0.1}$& 95.3$_{0.1}$ & 95.4$_{0.1}$ & \bf 95.7$_{0.2}$\\ %&  94.5$_{0.3}$\\
\bottomrule
\end{tabular}
% }
\caption{Results on different phases of adaptive pretraining compared to the baseline \roberta (col.~1).
Our approaches are \dapt (col.~2, \sect{sec:dapt}), \tapt (col.~3, \sect{sec:tapt}), and a combination of both (col.~4).
Reported results are development macro-$F_1$, except for \chemprot and \rct, for which we report micro-$F_1$, following \citet{Beltagy2019SciBERTAP}. We report averages across five random seeds, with standard deviations as subscripts.
$\dagger$ indicates high-resource settings. Best task performance is boldfaced. State-of-the-art results we can compare to: \chemprot (84.6), \rct (92.9), \arccite (71.0), \sciie (81.8), \hp (94.8), \ag (95.5), \imdb (96.2); references in \sect{sec:sota}. 
% \nascomment{is the boldface misplaced in the last row?}
}
\label{tab:dev_main}
\end{table*}
\begin{table*}[t]
\centering
\small
% \resizebox{\columnwidth}{!}{
\begin{tabular}{p{.7cm}p{1.5cm}rrr}
\toprule
%  & &  \multicolumn{2}{c}{Additional Pretraining}\\
% \cline{3-4}
\textbf{Dom.} &   
\textbf{Task} 
& {\bf \textsc{RoB.}} & \textbf{\dapt} & \bf $\lnot$\dapt\\
\midrule 
\multirow{2}{*}{\small \textsc{BM}} &   
\chemprot 
& 83.2$_{1.4}$& \bf 84.1$_{0.5}$ & 80.9$_{0.5}$\\
& 
$^\dagger$\rct  
& 88.1$_{0.0}$ & \bf 88.5$_{0.1}$ & 87.9$_{0.1}$\\ 
\midrule[0.03em]
\multirow{2}{*}{\small \cs} & 
{\footnotesize \arccite}
& 71.3$_{2.8}$& \bf 73.2$_{1.5}$ & 68.1$_{5.4}$ \\
& 
\sciie  
& 83.8$_{1.1}$& \bf 88.4$_{1.7}$& 83.9$_{0.9}$\\
\midrule[0.03em]
\multirow{2}{*}{\small \news}& 
\hpshort 
& \bf  84.0$_{1.5}$& 79.1$_{3.5}$& 71.6$_{4.6}$ \\
&$^\dagger$\ag 
& \bf 94.3$_{0.1}$& \bf  94.3$_{0.1}$ & 94.0$_{0.1}$ \\
\midrule[0.03em]
\multirow{2}{*}{\textsc{Rev.}} & 
$^\dagger$\textsc{Helpful}.
& 65.5$_{3.4}$& \bf 66.5$_{1.4}$&  65.5$_{3.0}$ \\
&$^\dagger$\imdb 
& 94.8$_{0.1}$& \bf  95.3$_{0.1}$ &  93.8$_{0.2}$\\
\bottomrule
\end{tabular}
% }
\caption{Development comparison of \roberta (\textsc{RoBa.}) and \dapt to adaptation to an \emph{irrelevant} domain ($\lnot$ \dapt). %For \news, we chose \cs as the irrelevant domain, for \reviews, we chose \med as the irrelevant domain, and for \cs and \med, we chose \news as the irrelevant domain. 
See \sect{sec:undomain} for our choice of irrelevant domains. Reported results follow the same format as Table \ref{tab:main}.
% \swabha{It would be nice to add the undomain domain to each row here...but it's a \latex nightmare to confine it to a single column.}
}
\label{tab:dev_undomain}
\end{table*}
\begin{table*}[ht]
\small
\centering
\resizebox{\textwidth}{!}{  
\begin{tabular}{l|ll}
\toprule
% \multirow{3}{*}{\rotatebox{90}{\textsc{BioMed}}}   
\med  
& \rct                             
& \chemprot  \\
%\cmidrule{2-4} 
\midrule
\tapt    
& 88.3$_{0.1}$ 
& 83.0$_{0.6}$  \\   
Transfer-\tapt 
& 88.0$_{0.1}$ ($\downarrow 0.3$) 
& 81.1$_{0.5}$ ($\downarrow 1.9$) \\

\bottomrule
\toprule
%  \multirow{3}{*}{\rotatebox{90}{\textsc{ReNews}}} 
\news        
 & \hp                     
 & \ag   \\    
%\cmidrule{2-4} 
\midrule
\tapt
& 82.7$_{3.3}$ 
& 94.7$_{0.1}$\\
Transfer-\tapt
& 77.6$_{3.6}$ ($\downarrow 5.1$) 
& 94.4$_{0.1}$ ($\downarrow 0.4$) \\
\bottomrule
\end{tabular}
\quad
\begin{tabular}{l|ll}
\toprule
%  \multirow{3}{*}{\rotatebox{90}{\textsc{CompSc}}} 
\cs   
 & \arccite                   
 & \sciie \\
%\cmidrule{2-4}  
\midrule
\tapt
& 73.2$_{3.6}$               
& 85.9$_{0.8}$ \\
Transfer-\tapt 
& 74.0$_{4.5}$ ($\uparrow 1.2$) 
& 85.5$_{1.1}$ ($\downarrow 0.4$) \\
\bottomrule
\toprule
% \multirow{3}{*}{\rotatebox{90}{\textsc{reviews}}} 
\amazon       
& \helpful                         
& \imdb     \\
%\cmidrule{2-4} 
\midrule
\tapt    
& 69.2$_{2.4}$ 
& 95.4$_{0.1}$\\
 Transfer-\tapt 
& 65.4$_{2.7}$ ($\downarrow 3.8$)                
& 94.9$_{0.1}$ ($\downarrow 0.5$) \\
\bottomrule
\end{tabular}}

\caption{Development results for \tapt transferability.}
\label{tab:transfer_dev_results}
\end{table*}
\begin{table*}[t]
\centering
\small
% \resizebox{\columnwidth}{!}{
\begin{tabular}{c|ccc}
\toprule
Pretraining &\med & \news & \reviews \\
                    & \rct-500          & \hp    & $^\dagger$\imdb         \\
\midrule

\tapt     & 80.5$_{1.3}$ & 82.7$_{3.3}$ & 95.4$_{0.1}$ \\
\dapt + \tapt & 83.9$_{0.3}$ & 80.8$_{2.3}$ & 95.7$_{0.2}$ \\
\midrule[0.03em]
\oracle  & 84.4$_{0.3}$ &  \bf\boldmath{84.9$_{1.9}$} & 95.8$_{0.1}$\\
\dapt + \oracle & \bf 84.5\boldmath{$_{0.3}$} & 83.1$_{3.7}$ & \bf96.0\boldmath{$_{0.1}$} \\
\bottomrule
\end{tabular}
% }
\caption{Mean development set macro-$F_1$ (for \hp and \imdb) and micro-$F_1$ (for \rct-500), with \oracle across five random seeds, with standard deviations as subscripts.
$\dagger$ indicates high-resource settings.}
\label{tab:dev_micro_vs_macro_task_adaptive_results}
\end{table*}
\begin{table*}[t]
\centering
\small
% \resizebox{\columnwidth}{!}{
\begin{tabular}{cccc}
\toprule
Pretraining  & \multicolumn{2}{c}{\med} & \cs  \\
&   \chemprot  & \rct-500         & \arccite  \\
\midrule
\roberta           &  83.2$_{1.4}$  &	80.3$_{0.5}$ &	71.3$_{2.8}$  \\
\tapt &    83.0$_{0.6}$ &	80.5$_{1.3}$ &	73.2$_{3.6}$   \\
\midrule[0.03em]
\random &  83.3$_{0.5}$  & 81.6$_{0.6}$	  &	 78.7$_{4.0}$     \\
\knn{50} &  83.3$_{0.8}$    &	81.7$_{0.5}$ &	70.1$_{3.5}$ \\
\knn{150} &   83.3$_{0.9}$       & 81.9$_{0.8}$ &  \bf 78.5$_{2.2}$ \\
\knn{500} &   \bf 84.5$_{0.4}$  &  82.6$_{0.4}$ &	77.4$_{2.3}$ \\
\midrule 
\dapt & 84.1$_{0.5}$ & \bf 83.5$_{0.8}$ & 73.2$_{1.5}$ \\
\bottomrule
\end{tabular}
% }
\caption{Mean development set macro-$F_1$ (for \hpshort and \imdb) and micro-$F_1$ (for \rct), across five random seeds, with standard deviations as subscripts, comparing \random (with 50 candidates) and \knn{k} selection.
Neighbors of the task data are selected from the domain data.
% Tried this notation to normalize random and nearest neighbors. \nascomment{see comment in text about abbreviations}
}
\label{tab:dev_selection_results}
\end{table*}
\begin{table*}
\centering
\small
\resizebox{\textwidth}{!}{
\begin{tabular}{>{\raggedright}p{9cm}p{9cm}}
\toprule 
\textbf{\imdb review}  & \textbf{\realnews article}\\
\midrule 
\textbf{Spooks} is enjoyable trash, featuring some well directed sequences, ridiculous plots and dialogue, and some third rate acting. Many have described this is a UK version of “24“, and one can see the similarities. The American version shares the weak silly plots, but the execution is so much slicker, sexier and I suspect, expensive. Some people describe weak comedy as “gentle comedy“. This is gentle spy story hour, the exact opposite of anything created by John Le Carre. Give me Smiley any day. &  [...] Remember poor Helen Flynn from \textbf{Spooks}? In 2002, the headlong BBC spy caper was in such a hurry to establish the high-wire stakes of its morally compromised world that Lisa Faulkner’s keen-as-mustard MI5 rookie turned out to be a lot more expendable than her prominent billing suggested. [...] Functioning as both a shocking twist and rather callous statement that No-One Is Safe, it gave the slick drama an instant patina of edginess while generating a record-breaking number of complaints.  [...] \\
\midrule
\textbf{The Sopranos} is perhaps the most mind-opening series you could possibly ever want to watch. It's smart, it's quirky, it's funny - and it carries the mafia genre so well that most people can't resist watching. The best aspect of this show is the overwhelming realism of the characters, set in the subterranean world of the New York crime families. For most of the time, you really don't know whether the wise guys will stab someone in the back, or buy them lunch. Further adding to the realistic approach of the characters in this show is the depth of their personalities - These are dangerous men, most of them murderers, but by God if you don't love them too. I've laughed at their wisecracks, been torn when they've made err in judgement, and felt scared at the sheer ruthlessness of a serious criminal. [...] & The drumbeat regarding the “Breaking Bad” finale has led to the inevitable speculation on whether the final chapter in this serialized gem will live up to the hype or disappoint (thank you, “Dexter,” for setting that bar pretty low), with debate, second-guessing and graduate-thesis-length analysis sure to follow. The Most Memorable TV Series Finales of All-Time [...] No ending in recent years has been more divisive than \textbf{“The Sopranos”} – for some, a brilliant flash (literally, in a way) of genius; for others (including yours truly), a too-cute copout, cryptically leaving its characters in perpetual limbo. The precedent to that would be “St. Elsewhere,” which irked many with its provocative, surreal notion that the whole series was, in fact, conjured in the mind of an autistic child. [...] \\
\midrule
\textbf{The Wicker Man}, starring Nicolas Cage, is by no means a good movie, but I can't really say it's one I regret watching. I could go on and on about the negative aspects of the movie, like the terrible acting and the lengthy scenes where Cage is looking for the girl, has a hallucination, followed by another hallucination, followed by a dream sequence- with a hallucination, etc., but it's just not worth dwelling on when it comes to a movie like this. Instead, here's five reasons why you SHOULD watch The Wicker Man, even though it's bad: 5. It's hard to deny that it has some genuinely creepy ideas to it, the only problem is in its cheesy, unintentionally funny execution. If nothing else, this is a movie that may inspire you to see the original 1973 film, or even read the short story on which it is based. 4. For a cheesy horror/thriller, it is really aesthetically pleasing. [...] NOTE: The Unrated version of the movie is the best to watch, and it's better to watch the Theatrical version just for its little added on epilogue, which features a cameo from James Franco. & [...] What did you ultimately feel about \textbf{''The Wicker Man''} movie when all was said and done? [...] I’m a fan of the original and I’m glad that I made the movie because they don’t make movies like that anymore and probably the result of what ''Wicker Man'' did is the reason why they don’t make movies like that anymore. Again, it’s kind of that ’70’s sensibility, but I’m trying to do things that are outside the box. Sometimes that means it’ll work and other times it won’t. Again though I’m going to try and learn from anything that I do. I think that it was a great cast, and Neil La Bute is one of the easiest directors that I’ve ever worked with. He really loves actors and he really gives you a relaxed feeling on the set, that you can achieve whatever it is that you’re trying to put together, but at the end of the day the frustration that I had with ‘The Wicker Man,’ which I think has been remedied on the DVD because I believe the DVD has the directors original cut, is that they cut the horror out of the horror film to try and get a PG-13 rating. I mean, I don’t know how to stop something like that. So I’m not happy with the way that the picture ended, but I’m happy with the spirit with which it was made. [...] \\
\midrule
Dr. Seuss would sure be mad right now if he was alive. \textbf{Cat in the Hat} proves to show how movie productions can take a classic story and turn it into a mindless pile of goop. We have Mike Myers as the infamous Cat in the Hat, big mistake! Myers proves he can't act in this film. He acts like a prissy show girl with a thousand tricks up his sleeve. The kids in this movie are all right, somewhere in between the lines of dull and annoying. The story is just like the original with a couple of tweaks and like most movies based on other stories, never tweak with the original story! Bringing in the evil neighbor Quin was a bad idea. He is a stupid villain that would never get anywhere in life. [...] & \textbf{The Cat in the Hat}, [...] Based on the book by Dr. Seuss [...] From the moment his tall, red-and-white-striped hat appears at their door, Sally and her brother know that the Cat in the Hat is the most mischievous cat they will ever meet. Suddenly the rainy afternoon is transformed by the Cat and his antics. Will their house ever be the same? Can the kids clean up before mom comes home? With some tricks (and a fish) and Thing Two and Thing One, with the Cat in The Hat, the fun's never done!Dr. Seuss is known worldwide as the imaginative master of children's literature. His books include a wonderful blend of invented and actual words, and his rhymes have helped many children and adults learn and better their understanding of the English language. [...] \\
\bottomrule
\end{tabular}}
\caption{Additional examples that highlight the overlap between \imdb reviews and \realnews articles.}
\label{tab:more-domain-overlap}
\end{table*}
%  \begin{figure*}[t]
%       \centering
%       \includegraphics[width=\textwidth]{figs/selection_fig_ana.png}
%       \caption{An illustration of automated data selection for TAPT with VAMPIRE. 
%       } 
%       \label{fig:selection_intuition}
% \end{figure*}

\begin{table*}
\centering
\small
 \resizebox{\textwidth}{!}{
\begin{tabular}{p{1.5cm}p{15cm}}
\toprule 
%   & \textbf{\realnews article}\\
Source& During median follow-up of 905 days ( IQR 773-1050 ) , 49 people died and 987 unplanned  admissions were recorded ( totalling 5530 days in hospital ) .\\
\midrule[0.03em]
Neighbor 0& Of this group, 26\% died after discharge from hospital, and the median time to death was 11 days (interquartile range, 4.0-15.0 days) after discharge.\\
Neighbor 1& The median hospital stay was 17 days (range 8-26 days), and all the patients were discharged within 1 month. \\
Neighbor 2& The median hospital stay was 17 days (range 8-26 days).\\
Neighbor 3& The median time between discharge and death was 25 days (mean, 59.1 days) and no patient was alive after 193 days.\\
Neighbor 4& The length of hospital stay after colostomy formation ranged from 3 days to 14 days with a median duration of 6 days (+IQR of 4 to 8 days).\\
\midrule
Source& Randomized , controlled , parallel clinical trial .\\
\midrule[0.03em]
Neighbor 0& Design: Unblinded, randomised clinical controlled trial.\\
Neighbor 1& These studies and others led to the phase III randomized trial RTOG 0617/NCCTG 0628/ CALGB 30609.\\
Neighbor 2& -Definitive randomized controlled clinical trial (RCT):\\
Neighbor 3& RCT $\frac{1}{4}$ randomized controlled trial.\\
Neighbor 4& randomized controlled trial [ Fig. 3(A)].\\
\midrule
Source& Forty primary molar teeth in 40 healthy children aged 5-9 years were treated by direct pulp capping .\\
\midrule[0.03em]
Neighbor 0& In our study, we specifically determined the usefulness of the Er:YAG laser in caries removal and cavity preparation of primary and young permanent teeth in children ages 4 to 1 8 years.\\
Neighbor 1& Males watched more TV than females, although it was only in primary school-aged children and on weekdays.\\
Neighbor 2& Assent was obtained from children and adolescents aged 7-17 years.\\
Neighbor 3& Cardiopulmonary resuscitation was not applied to children aged <5 years (Table 2).\\
Neighbor 4& It measures HRQoL in children and adolescents aged 2 to 25 years.\\
\bottomrule
\end{tabular}}
% }
\caption{5 nearest neighbors of sentences from the \rct dataset (Source) in the \med domain (Neighbors 0--4).} 
\label{tab:knn_ex}
\end{table*}

\end{document}